\documentclass{article}

    \PassOptionsToPackage{numbers, compress}{natbib}

    \usepackage[preprint]{neurips_2024}

\usepackage[utf8]{inputenc} 
\usepackage[T1]{fontenc}    
\usepackage{hyperref}       
\usepackage{url}            
\usepackage{booktabs}       
\usepackage{amsfonts}       
\usepackage{nicefrac}       
\usepackage{microtype}      
\usepackage[table]{xcolor}         

\usepackage{enumitem}
\usepackage{amsmath}
\usepackage{multirow}
\usepackage{adjustbox}
\usepackage{array}
\usepackage{booktabs} 
\usepackage{tabularx}
\usepackage{graphicx}
\usepackage{caption}

\usepackage{lipsum}
\newcommand\blfootnote[1]{%
\begingroup
\renewcommand\thefootnote{}\footnote{#1}%
\addtocounter{footnote}{-1}%
\endgroup
}

\usepackage[most]{tcolorbox}

\usepackage{booktabs}
\usepackage{longtable}
\usepackage{geometry}
\definecolor{lightgray}{gray}{0.95}
\definecolor{lightblue}{RGB}{230,245,255}
\definecolor{lightgreen}{RGB}{240,255,240}

\definecolor{codegreen}{rgb}{0,0.6,0}
\definecolor{codegray}{rgb}{0.5,0.5,0.5}
\definecolor{codepurple}{rgb}{0.58,0,0.82}
\definecolor{backcolour}{rgb}{0.95,0.95,0.92}
\usepackage{listings}
\lstdefinestyle{mystyle}{
  commentstyle=\color{codegreen},
  keywordstyle=\color{magenta},
  numberstyle=\small\color{codegray},
  stringstyle=\color{codepurple},
  basicstyle=\small,
  breakatwhitespace=false,         
  breaklines=true,                 
  captionpos=b,                    
  keepspaces=false,                                 
  showspaces=false,                
  showstringspaces=false,
  showtabs=false,                  
  tabsize=2
}

\title{WebGen-Bench: Evaluating LLMs on Generating Interactive and Functional Websites from Scratch}

\author{%
  Zimu Lu$^{*}$, Yunqiao Yang$^{*}$, Houxing Ren$^{*\S}$, Han Xiao, Ke Wang, Weikang Shi \\ {\bf Aojun Zhou}, {\bf Mingjie Zhan}$^{\dagger}$, {\bf Hongsheng Li}$^{\dagger}$\\
  Multimedia Laboratory (MMLab), The Chinese University of Hong Kong\\
 \texttt{luzimu@mail.ustc.edu.cn} \quad \texttt{hsli@ee.cuhk.edu.hk}
}

\begin{document}

\maketitle

\begin{abstract}

LLM‑based agents have demonstrated great potential in generating and managing code within complex codebases. In this paper, we introduce WebGen-Bench, a novel benchmark designed to measure an LLM-based agent's ability to create multi-file website codebases from scratch. It contains diverse instructions for website generation, created through the combined efforts of human annotators and GPT-4o. These instructions span three major categories and thirteen minor categories, encompassing nearly all important types of web applications. To assess the quality of the generated websites, we use GPT-4o to generate test cases targeting each functionality described in the instructions, and then manually filter, adjust, and organize them to ensure accuracy, resulting in 647 test cases. Each test case specifies an operation to be performed on the website and the expected result after the operation. To automate testing and improve reproducibility, we employ a powerful web-navigation agent to execute tests on the generated websites and determine whether the observed responses align with the expected results. We evaluate three high-performance code-agent frameworks—Bolt.diy, OpenHands, and Aider—using multiple proprietary and open-source LLMs as engines. The best-performing combination, Bolt.diy powered by DeepSeek-R1, achieves only 27.8\% accuracy on the test cases, highlighting the challenging nature of our benchmark. Additionally, we construct WebGen-Instruct, a training set consisting of 6,667 website-generation instructions. Training Qwen2.5-Coder-32B-Instruct on Bolt.diy trajectories generated from a subset of this training set achieves an accuracy of 38.2\%, surpassing the performance of the best proprietary model. We release our data-generation, training, and testing code, along with both the datasets and model weights at~\url{https://github.com/mnluzimu/WebGen-Bench}, to facilitate future research.

\end{abstract}
\blfootnote{$^*$Equal contribution\quad $^\dagger$Corresponding author\quad $^\S$Project lead}

\section{Introduction}

Recent developments in large language models (LLMs) have demonstrated increasingly strong performance. When paired with agent frameworks, they have become much more competent at solving challenging tasks such as fixing bugs in complex codebases and competing in coding competitions. Prior works have sought to quantify the software engineering abilities of these LLM-powered agents by testing them on curated GitHub issues~\citep{jimenez2023swe, yang2024swe} and feature-patching requests~\citep{miserendino2025swe}. These tasks involve advanced modifications to existing codebases and primarily target expert engineers.

\begin{table}[t]\fontsize{9}{11}\selectfont
\centering
\caption{Comparison of WebGen-Bench with other repository-level software engineering benchmarks. $^*$ indicates that the statistics for SWE-Bench Multimodal are median values, whereas the others are average values. The values for our benchmarks are gathered from the test results of Bolt.diy, OpenHands, and Aider using DeepSeek-V3. The values for the other benchmarks are taken from~\cite{jimenez2023swe},~\cite{miserendino2025swe}, and~\cite{yang2024swe}, respectively.}
\begin{tabularx}{\columnwidth}{>{\raggedright\arraybackslash\hsize=1.3\hsize}X |>{\centering\arraybackslash\hsize=.7\hsize}X >{\centering\arraybackslash\hsize=.7\hsize}X >{\centering\arraybackslash\hsize=.8\hsize}X >{\centering\arraybackslash\hsize=.7\hsize}X}
\toprule
\textbf{Benchmark} & \textbf{From Scratch} & \textbf{Training Set} & \textbf{Number of Files} & \textbf{Lines of Code} \\
\midrule
WebGen-Bench (ours) & \textcolor{green}{\checkmark} & \textcolor{green}{\checkmark} & 8.1 & 315.3 \\
SWE-Bench & \textcolor{red}{\texttimes} & \textcolor{green}{\checkmark} & 1.7 & 32.8 \\
SWE-Bench Multimodal$^*$ & \textcolor{red}{\texttimes} & \textcolor{red}{\texttimes} & 2 & 27 \\
SWE-Lancer & \textcolor{red}{\texttimes} & \textcolor{red}{\texttimes} & 2 & 55 \\
\bottomrule
\end{tabularx}
\label{tab:web_bench_comparison}
\end{table}

On the other hand, there is a growing need for code agents to assist non-experts with little or no programming background in building applications tailored to their needs and expectations. For example, Bolt.new\footnote{https://bolt.new} and Lovable.dev\footnote{https://lovable.dev} are two projects that generate complete websites based on user requests and have become very popular among customers. This task poses significant challenges for LLM-based agents, as building a fully functional and customized web application from scratch tests a wide range of capabilities—including high-level planning, organizing complex multi-file codebases, and implementing nuanced user requirements in both functionality and design. However, there is currently a lack of systematic and reliable evaluation methods for this task. The high demand for such applications, coupled with the value of assessing agent capabilities, highlights the need for a novel benchmark to evaluate the ability to generate websites from scratch based on natural language instructions.

To this end, we introduce \textbf{WebGen‑Bench}, the first benchmark to systematically evaluate LLM‑based agents' ability to construct websites that satisfy the functional and appearance requirements specified in user instructions. As shown in Table~\ref{tab:web_bench_comparison}, unlike prior software‑engineering benchmarks \cite{jimenez2023swe,miserendino2025swe,yang2024swe}, which focus on fixing bugs or supplying patches to existing codebases, our benchmark requires models to build a complex codebase from scratch, assessing agents' ability to plan, develop, and manage projects with multi‑file structures. There are two critical challenges to address when creating the benchmark: (1) how to curate diverse instructions covering major web‑application categories and (2) how to accurately evaluate the websites generated from scratch. WebGen‑Bench tackles these problems by providing a systematic data‑curation process and a reliable evaluation pipeline that thoroughly assesses the agents' website‑generation abilities.

To construct the benchmark dataset, we initially examined prominent platforms listing website‑development projects, such as Upwork\footnote{https://www.upwork.com/}, Freelancer\footnote{https://www.freelancer.com/}, and Proginn\footnote{https://www.proginn.com}. Through systematic analysis and discussions among the authors, we identified twenty common categories of website‑development requirements. Subsequently, a panel of forty computer‑science Ph.D. students conducted extensive brainstorming sessions to determine numerous specific web applications for each category, along with brief, clear functionality and appearance requirements for each application. Then, we employed GPT‑4o to convert these web applications and their corresponding requirements into diverse website‑generation instructions. Next, we prompted GPT‑4o to generate test cases for each instruction, targeting both functional and appearance requirements. Each test case consists of an operation to be performed on the website and the expected outcome. Finally, we manually reviewed each test case, verifying that it is reasonable and comprehensively covers all requirements specified in the corresponding instruction.

\begin{figure*}[t]
    \centering
    \includegraphics[width=1.0\textwidth]{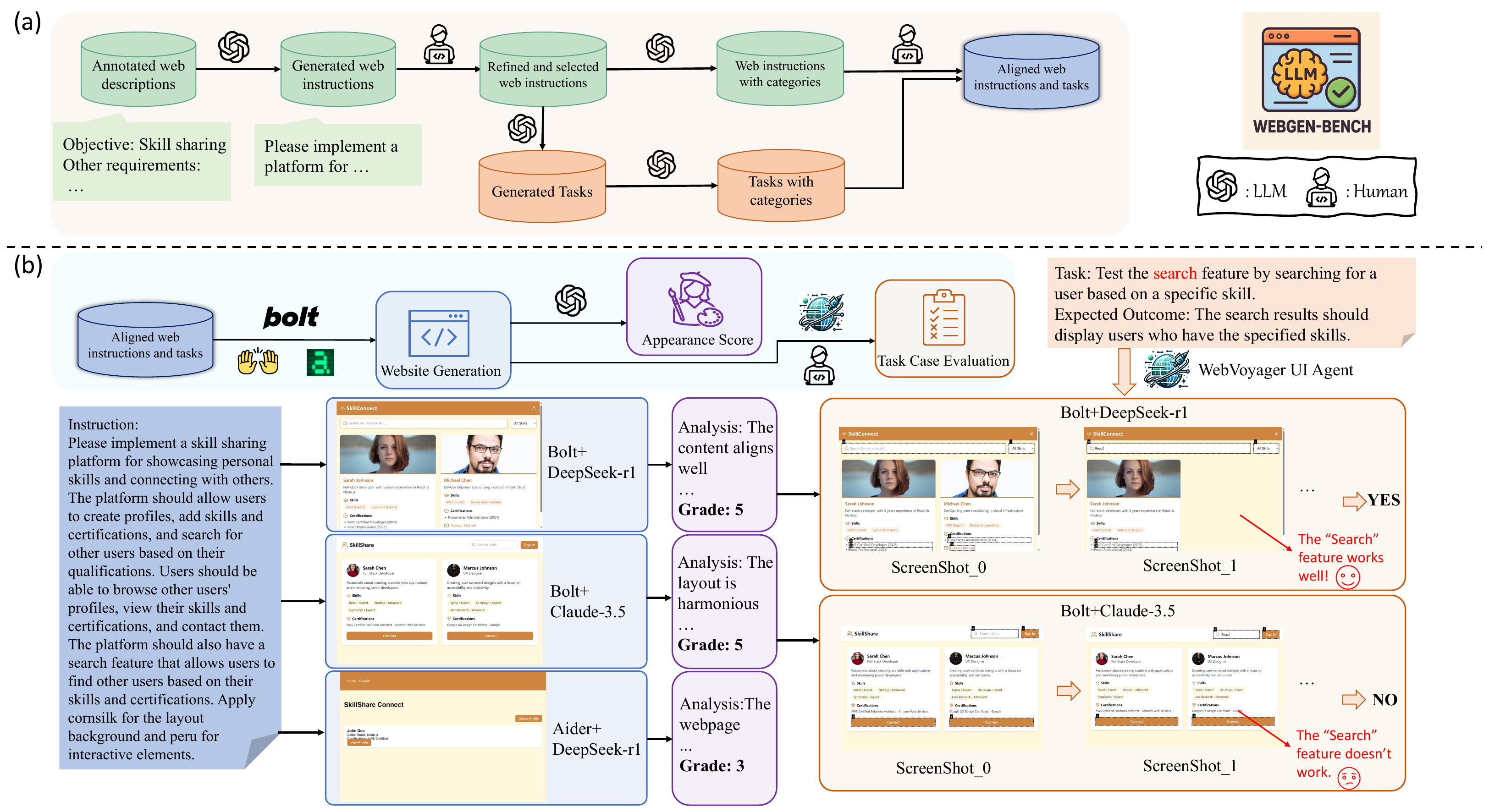}
    \caption{The data‑curation and automatic‑testing pipeline. (a) depicts the process for curating website‑generation instructions together with their corresponding test cases. (b) presents the testing pipeline for verifying whether the generated websites meet functionality and design requirements with the WebVoyager UI agent, and for analyzing their aesthetic quality using GPT‑4o.}
    
\label{fig:data_and_testing_pipeline}
\end{figure*}

Using human testers for evaluation is slow, costly, and presents management challenges. Inspired by~\cite{lu2025uxagent}, which employs agents with specified personas for automated testing, we utilize a strong web navigation agent, WebVoyager~\cite{he2024webvoyager}, to perform operations and assess outcomes. Furthermore, we also evaluates the harmony and aesthetic quality of the website design by designing a prompt that asks GPT-4o to grade the appearance of the generated websites on a scale from 1 to 5. This provides a comprehensive assessment of overall website generation quality. The testing process is demonstrated in Fig.~\ref{fig:data_and_testing_pipeline} (b). With this testing framework, we evaluate the performance of three strong open-source code agents — Bolt.diy, OpenHands, and Aider — using DeepSeek-V3 as the engine LLM. Among them, Bolt.diy achieves the best performance. We further evaluate Bolt.diy across a diverse set of proprietary and open-source models. DeepSeek-R1 achieves the best performance, attaining an accuracy of only 27.8\% on the test cases, while Claude-3.5-Sonnet achieves the highest average appearance score of 3.0. The low accuracies and appearance scores demonstrate that the performance of current LLM-based agents remains far from saturated on our benchmark.

We also construct a training dataset named \textbf{WebGen‑Instruct}, which contains 6,667 website‑generation instructions. To avoid data contamination, we removed instructions that are semantically similar to those in WebGen‑Bench by applying Jaccard‑similarity filtering and Sentence‑Transformers–based deduplication~\citep{reimers-2019-sentence-bert}, as detailed in Appendix~\ref{sec:decontamination}. Fine‑tuning Qwen2.5‑Coder‑32B‑Instruct on Bolt.diy trajectories—generated from a subset of WebGen‑Instruct by DeepSeek‑V3 with rejection sampling raises its accuracy to 38.2\%, a substantial improvement over its original 9.5\% and even higher than the performance of DeepSeek‑R1. We also fine‑tune Qwen2.5‑Coder‑7B‑Instruct and Qwen2.5‑Coder‑14B‑Instruct on the same training data, and name the resulting family of website‑generation models \textbf{WebGen‑LM}.

Our contributions are as follows:

\begin{itemize}[leftmargin=*] 
    \item We introduce WebGen-Bench, the first benchmark designed to test the ability of an LLM-based agent to generate websites from scratch. It includes diverse instructions for website generation and corresponding test cases to evaluate website functionalities. 
    \item We conduct comprehensive evaluations of three high-performance code-agent frameworks — Bolt.diy, OpenHands, and Aider — using different proprietary LLMs as engines, demonstrating the challenging nature of our benchmark.
    \item We construct WebGen-Instruct, a training set consisting of 6,667 website-generation instructions. We use this training set to fine-tune Qwen2.5-Coder-Instruct models of sizes 7B, 14B, and 32B, resulting in a family of LLMs specialized in website generation, named WebGen-LM. WebGen-LM-32B achieves an accuracy of 38.2\% on WebGen-Bench, surpassing DeepSeek-R1. 
\end{itemize}

\section{Related Work}

\textbf{Software Engineering Benchmarks.} Code generation has long been used as a means to evaluate the abilities of LLMs~\citep{hendrycks2021measuring,chen2021evaluating,austin2021program}. Previous works have collected coding problems from various sources, such as user queries~\citep{zhang2024naturalcodebench}, coding contests~\citep{jain2024livecodebench}, model synthesis~\citep{zhuo2024bigcodebench}, and expert design~\citep{muennighoff2023octopack,chen2021evaluating}, to evaluate LLMs' performance on single-file, function-level coding tasks.
Recently, as stronger models have reached a plateau on these simpler benchmarks, new benchmarks such as SWE-bench~\citep{jimenez2023swe,yang2024swe} and SWE-Lancer~\citep{miserendino2025swe} have been constructed by collecting real-world code repositories and corresponding issue requests to test models' ability to solve bugs and implement new functionalities.
These benchmarks require models to identify and fix issues~\citep{jimenez2023swe,yang2024swe,aleithan2024swe}, perform code completions~\citep{liu2023repobench,zhang2023repocoder}, or provide functionality patches~\citep{miserendino2025swe} within an existing multi-file codebase.
Different from previous works, our benchmark focuses on creating web applications from scratch based on natural language instructions, requiring models to generate a complex, multi-file codebase, implement multiple functionality and appearance requirements, and make independent technical design decisions.

\textbf{LLM-based Code Agents and Pipelines.} Various agent-based~\citep{wang2024openhands,yang2024sweagent,aiderai2024aider,Cursor2024cursor,GitHub2024copilot,Wu2024devin} and pipeline-based~\citep{xia2024agentless,ruan2024specrover,zhang2024autocoderover} methods have been proposed to address software engineering problems such as code completion and GitHub issue resolution. While pipeline-based methods sometimes demonstrate strong performance on specific tasks with fixed pipelines~\citep{jimenez2023swe}, agent-based methods are generally more flexible and versatile. Code agent frameworks such as OpenHands~\citep{wang2024openhands} and SWE-agent~\citep{yang2024sweagent} interact with executable environments to obtain feedback from the execution of generated code. To evaluate our benchmark, we selected three open-source code agents. Among them, OpenHands~\citep{wang2024openhands} and Aider~\citep{aiderai2024aider} are general-purpose code agent frameworks that we adapted for our benchmark, while Bolt.diy~\citep{stackblitzlabs2024bolt} is a specialized framework for generating web applications. Prior works~\citep{pan2024training,ma2025sorft,ma2024lingma,xie2025swe,ma2024repository} have employed various post-training methods to improve the performance of code agents using open-source models. In this work, we also fine-tune open-source models with trajectories generated from web-generating instructions.

\textbf{Automatic Software User-testing.} User-testing is a common method in software engineering to assess the functionality of software with high user-interaction requirements. However, human testing can be costly and introduce significant management complexities. Various works have employed agents to test websites~\cite{lu2025uxagent}, graphical user interfaces (GUIs)~\citep{9196452}, and games~\citep{10.1145/3290607.3313039,9440143}. Among them, UXAgent~\citep{lu2025uxagent} uses UI agents with pre-defined personas to simulate user experiences on websites. Our work also utilizes a web navigation UI agent to evaluate generated websites. Different from prior works, we define atomic test cases targeting functionality and appearance requirements, enabling the agent to perform operations and observe whether the website behaves as intended.

\section{WebGen-Bench}

In this section, we introduce WebGen-Bench, the first benchmark designed to test the ability of LLM-based agents to generate websites from scratch based on natural language instructions. The benchmark consists of diverse website-generation instructions and comprehensive test cases that have been carefully constructed and repeatedly validated. A reliable and cost-effective testing pipeline, built around a strong web navigation agent, has been developed to ensure efficient evaluation of the generated websites. The data curation process and testing pipeline are shown in Fig.~\ref{fig:data_and_testing_pipeline} (a) and (b) respectively.

\subsection{Instruction Curation}

\textbf{Web Development Project Descriptions Collection.} To ensure the diversity and practicality of the instructions, we first carefully browsed several platforms containing website development project listings, including Upwork\footnote{https://www.upwork.com}, Freelancer\footnote{https://www.freelancer.com}, and Proginn\footnote{https://www.proginn.com}. After systematic analysis and discussions among the authors, we identified twenty prevalent web application categories, as outlined in Table~\ref{tab:category_number}. To simulate numerous customized web applications, we employ a panel of forty computer science Ph.D students to conduct brainstorming sessions to determine various specific web applications belonging to these categories, as well as a brief and clear list of corresponding functionality and appearance requirements for each application. A customized application and its corresponding requirements are combined into a project description. We manually created 10152 project descriptions in total.

\textbf{Web Development Project Descriptions Collection.} To ensure the diversity and practicality of the instructions, we first carefully browsed several platforms containing website development project listings, including Upwork\footnote{https://www.upwork.com}, Freelancer\footnote{https://www.freelancer.com}, and Proginn\footnote{https://www.proginn.com}. After systematic analysis and discussions among the authors, we identified twenty prevalent web application categories, as outlined in Table~\ref{tab:category_number}. To simulate numerous customized web applications, we employ a panel of forty computer science Ph.D students to conduct brainstorming sessions to determine various specific web applications belonging to these categories, as well as a brief and clear list of corresponding functionality and appearance requirements for each application. A customized application and its corresponding requirements are combined into a project description. We manually created 10152 project descriptions in total.

\textbf{Website‑Generation Instruction Curation.}
From the collected project descriptions, we use one‑shot prompting with GPT‑4o to generate the corresponding instructions. The prompt template is shown in Fig.~\ref{fig:instruction_generation} in Appendix~\ref{sec:prompt_derive_instruction}. Because the total number of generated instructions exceeds the practical limits of benchmarking code agents—which require substantial computational resources and long inference trajectories—we sample two to eight representative examples from each category to preserve both coverage and diversity. This procedure produces a curated test set containing 101 instructions.

Next, we deduplicate the remaining instructions by first filtering those with a 5-gram Jaccard similarity score exceeding 0.6 relative to any testing instruction. We then perform semantic deduplication by computing cosine similarity between sentence embeddings~\citep{reimers-2019-sentence-bert} of the remaining instructions and the testing set. This decontamination process produces a training set of 6,667 website-generation instructions, which we name \textbf{WebGen-Instruct}. Details of the decontamination process are provided in Appendix~\ref{sec:decontamination}.

\textbf{Test Set Adjustment and Validation.} We refine and validate the selected test instructions to ensure they exclude unreasonable designs and specific technical implementation details. We intentionally omit technical design specifications because our dataset aims to evaluate code agents in scenarios where they receive instructions from non-expert users. The agents should autonomously determine the optimal technical approach. Including tool-specific hints in the instructions would compromise this objective.

\textbf{Technical Classification of the Testing Set.} Given the limited number of testing instructions per application category, analyzing categorical statistics based on the original 20 application categories would be confusing. To enable higher-level analysis, we reorganize the 101 testing instructions into three broader technical categories (see Tab.~\ref{tab:bench_category_number}:)
1. Content Presentation: Static page generation (e.g., corporate/portfolio sites), dynamic rendering (e.g., blogs/news feeds), data visualization (e.g., dashboards), and immersive media displays (e.g., 360° product views).
2. User Interaction: Form systems, authentication flows, real-time collaboration tools, e-commerce transactions, and AI-enhanced features (e.g., chatbots).
3. Data Management: CRUD operations for content administration, third-party API integrations (e.g., payment/social platforms), analytical processing of user behavior data, and file operations (e.g., cloud synchronization, bulk exports).

\begin{table}[t]
\fontsize{9}{10}\selectfont
\centering
\caption{The number of website-generation instructions in each technical category in WebGen-Bench is shown. Each main category contains multiple subcategories. A sample may belong to one main category and multiple subcategories.}
\label{tab:bench_category_number}
\begin{tabularx}{\columnwidth}{>{\raggedright\arraybackslash\hsize=0.6\hsize}X >{\centering\arraybackslash\hsize=0.4\hsize}X | > {\raggedright\arraybackslash\hsize=0.6\hsize}X >{\centering\arraybackslash\hsize=0.4\hsize}X}
\toprule
\textbf{Main Categories}  & \textbf{Sample Number} & \textbf{Sub Category} & \textbf{Sample Number}  \\
\midrule
\multirow{4}*{Content Presentation} & \multirow{4}*{28} & Static Page Generation & 20 \\
& & Dynamic Content Rendering & 18  \\
& & Data Visualization & 36  \\
& &Media Display & 6 \\
\cmidrule(lr){1-4}
\multirow{5}*{User Interaction} & \multirow{5}*{49} & Form Systems & 40 \\
& & Authentication & 18  \\
& & Real-time Features & 20  \\
& &E-commerce & 22 \\
& &AI Integration & 19 \\
\cmidrule(lr){1-4}
\multirow{4}*{Data Management} & \multirow{4}*{24} & CRUD Operations & 29 \\
& & API Integration & 20  \\
& & Big Data & 12  \\
& &File Handling & 5 \\
\cmidrule(lr){1-4}
Total & 101 & & \\
\bottomrule
\end{tabularx}
\end{table}

\subsection{Test Case Construction and Evaluation}

Since the websites are generated from scratch based on the instructions, the tested agents have significant freedom in their implementation choices. To accurately evaluate how well the agents satisfy the instruction requirements while accommodating diverse implementation approaches, we construct test cases targeting each and every requirement in the instructions.

\textbf{Test Case Construction.} Each test case consists of an operation verifying a specific functionality or appearance requirement, paired with its expected outcome. We first generate draft test cases using GPT-4o with the prompt shown in Figure~\ref{fig:test_case_construction}. Two computer science Ph.D. students then independently review and refine these test cases. After comparing their adjustments, we resolved discrepancies through discussion, yielding a final set of 647 test cases (4–11 per instruction). This manual validation process guarantees strict alignment between test cases and instructions, ensuring: (1) all instruction requirements are covered by test cases, and (2) each test case corresponds to an instruction requirement. This approach ensures comprehensive evaluation while preserving implementation flexibility for the tested agents.

\textbf{UI Agent-based Evaluation.}
With instructions and test cases prepared, we must determine how to effectively evaluate the generated websites. Manual testing by human evaluators is costly and time-consuming, as completing a test case takes at least 60 seconds, and finishing all 647 test cases would require approximately 10.8 hours at an estimated cost of \$377.8~\citep{uswage}. This slow, labor-intensive process would hinder rapid iteration during framework development, preventing researchers from obtaining timely feedback when refining website-generation systems.

To improve testing efficiency, we automate test case evaluation. Inspired by~\cite{lu2025uxagent}, which employs persona-based agents for web usability testing, we utilize WebVoyager~\cite{he2024webvoyager}, a robust web navigation UI agent, to execute test operations and verify outcomes. We structure each test case's operation and expected outcome into a standardized prompt (Fig.~\ref{fig:agent_starting_prompt}), which directs the agent to simulate user interactions, analyze action trajectories and screenshots, and return YES, NO, or PARTIAL assessments based on requirement fulfillment. The process is shown on the right side of Fig.~\ref{fig:data_and_testing_pipeline} (b). When the agent reaches its interaction limit without completing evaluation, we trigger a decision prompt, inducing the agent to make a final decision (Fig.~\ref{fig:limit_reached_prompt}). Considering the cost induced by multiple interactions with the website in evaluating each test case, we employ Qwen2.5-VL-32B-Instruct, an efficient open-source vision-language model that balances performance and cost-effectiveness, as the agent's engine.

\subsection{Evaluation of Website Appearance}

\begin{figure*}[t]
    \centering
    \includegraphics[width=0.75\textwidth]{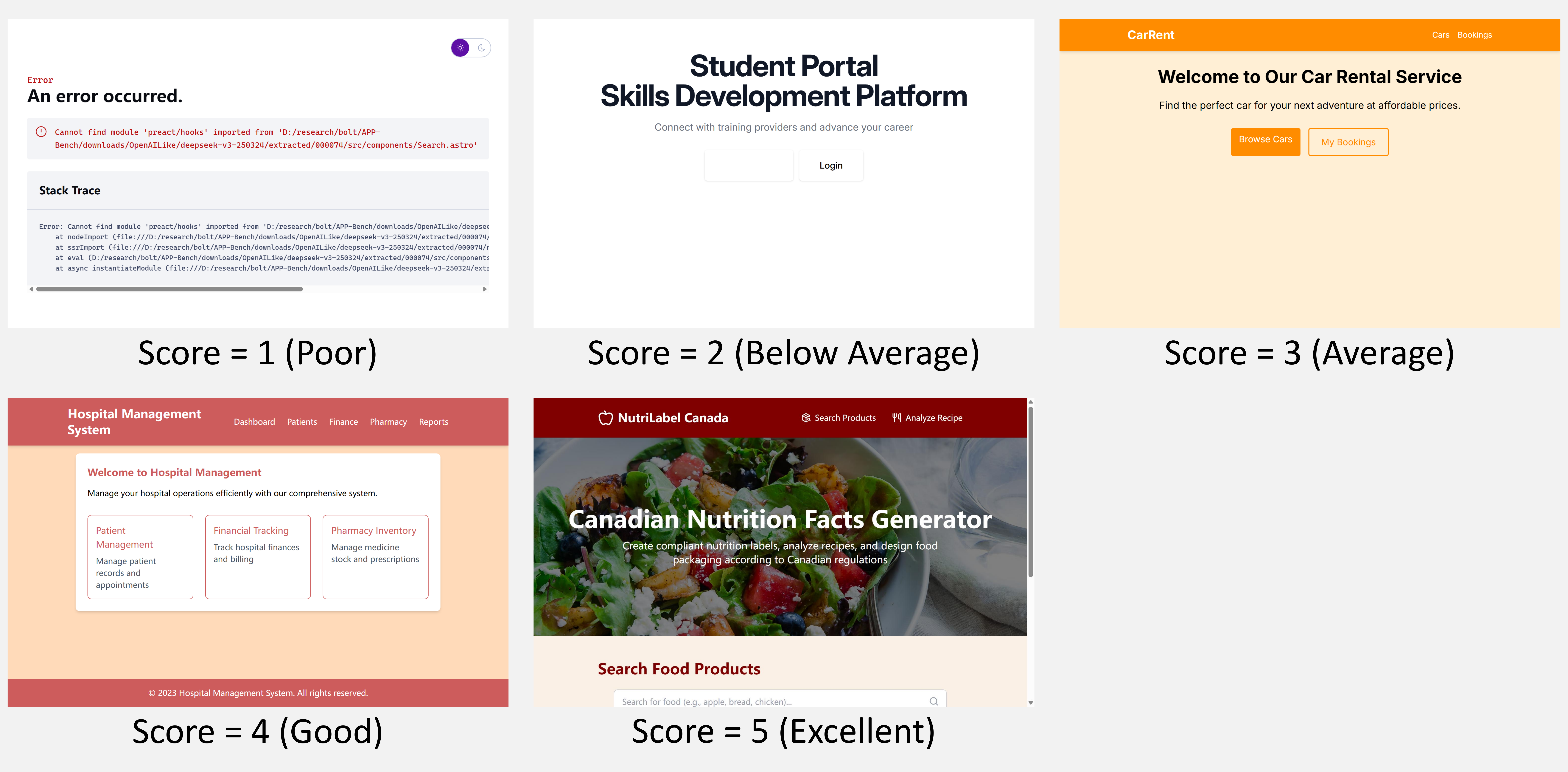}
    \caption{The examples of website screenshots at each appearance score.}
    
\label{fig:appearance}
\end{figure*}

Apart from the fulfillment of the functionality and appearance constraints in the instructions, another important aspect of website generation is the level of relevance, harmony, and aesthetics of the webpage. To conduct a quantitative analysis of this aspect, we designed a set of detailed metrics, ranging from the success of rendering and the relevance of the content to the harmony of the layout and the modernness of the design. We then place the metrics in a prompt, asking GPT-4o to grade the appearance of the website with a score ranging from 1 to 5 (the higher the better), as demonstrated in the middle part of Fig.~\ref{fig:data_and_testing_pipeline} (b).
The prompt is shown in Fig.~\ref{fig:appearance_grading_prompt}. Examples of websites at different appearance scores are shown in Fig.~\ref{fig:appearance}, and more examples are presented in Fig.~\ref{fig:appearance_scores} of Appendix.~\ref{sec:appearance_score_examples}.

\section{Experiments}



\subsection{Experimental Setup}

\textbf{Frameworks.} We evaluate three popular code-agent frameworks: Bolt.diy~\citep{stackblitzlabs2024bolt}, OpenHands~\citep{wang2024openhands}, and Aider~\citep{aiderai2024aider}. Bolt.diy is the open-source version of Bolt.new\footnote{https://bolt.new}, a browser-based framework for generating and previewing web applications. It provides a user interface and a Linux-like WebContainer environment that can execute code. It first prompts the model to decide which frontend and backend frameworks to use (such as Vite, React, Remix, etc.), then imports the basic template and builds upon it. OpenHands is a platform for AI-powered software development agents. For OpenHands, we pair it with CodeActAgent to evaluate it on our benchmark. The adapted instruction is presented in Appendix~\ref{sec:openhands_prompt}. Aider is a terminal-based AI programming framework that natively supports many popular programming languages, including Python, JavaScript, PHP, HTML, CSS, and more. Aider constructs a map of the entire codebase, which helps it function well in larger projects. We use the adapted instruction in Appendix~\ref{sec:aider_prompt} to generate websites with Aider.

\textbf{Models.} We first evaluate the three frameworks on DeepSeek-V3~\citep{liu2024deepseek}, Claude-3.5-Sonnet~\citep{anthropic2024claude}, and DeepSeek-R1~\citep{guo2025deepseek}. We then evaluate several strong general-purpose proprietary and open-source LLMs—including Claude-3.5-Sonnet~\citep{anthropic2024claude}, DeepSeek-R1~\citep{guo2025deepseek}, GPT-4o~\citep{hurst2024gpt}, O3-mini~\citep{openai2025o3}, Qwen2.5-Coder-32B~\citep{hui2024qwen2}, and Qwen2.5-72B-Instruct~\citep{yang2024qwen2}—on the best-performing framework, Bolt.diy. We do not test general-purpose models smaller than Qwen2.5-Coder-32B, as we observe that such models often fail to follow the specified output format and therefore cannot generate valid websites.

\textbf{Training Details.} To validate the effectiveness of our training set, we selectively generated Bolt.diy trajectories for a subset of 2K instructions from WebGen-Instruct using DeepSeek-V3. Using rejection sampling~\citep{yuan2023scaling}, we retained only the trajectories whose corresponding websites achieved an appearance score greater than or equal to 3, resulting in 600 trajectories. This filtering ensures that the remaining generated websites are relevant to the instructions and do not exhibit major rendering issues. We then fine-tuned Qwen2.5-Coder-Instruct models of sizes 7B, 14B, and 32B for 2 epochs, with a learning rate of 4e-5 and a batch size of 32. The 7B, 14B, and 32B models were trained on 8, 16, and 32 A800 GPUs, respectively. This fine-tuning process yields a family of models specialized in website generation, which we name \textbf{WebGen-LM}.

\subsection{Experimental Results}

\begin{table}[t]\fontsize{9}{10}\selectfont
\centering
\caption{Evaluation of three powerful code-agent frameworks using different proprietary and open-source models. Accuracy is computed using a weighted score, where YES samples are weighted by 1 and PARTIAL samples are weighted by 0.5; the total score is then divided by the number of test cases. The highest accuracy and appearance scores are marked in \textbf{bold}.}
\begin{tabularx}{\textwidth}{>{\raggedright\arraybackslash\hsize=1.4\hsize}X>{\centering\arraybackslash\hsize=0.6\hsize}X >{\centering\arraybackslash\hsize=0.6\hsize}X >{\centering\arraybackslash\hsize=0.6\hsize}X >{\centering\arraybackslash\hsize=0.7\hsize}X >{\centering\arraybackslash\hsize=0.6\hsize}X >{\centering\arraybackslash\hsize=0.6\hsize}X}
\toprule
\textbf{Test Name} & \textbf{Yes Rate} & \textbf{Partial Rate} & \textbf{No Rate} & \textbf{Start Failed} & \textbf{Accuracy} & \textbf{Appearance Score} \\
\midrule
\textbf{Bolt.diy} \\
\midrule
Claude-3.5-Sonnet & 22.6 & 7.6 & 64.1 & 5.7 & 26.4 & \textbf{3.0} \\
DeepSeek-R1 & 24.7 & 6.2 & 64.3 & 4.8 & 27.8 & 2.5 \\
DeepSeek-V3 & 18.5 & 4.5 & 73.9 & 3.1 & 20.8 & 2.0 \\
GPT-4o & 10.4 & 4.8 & 64.5 & 20.4 & 12.8 & 1.5 \\
o3-mini & 17.9 & 3.4 & 40.0 & 38.6 & 19.6 & 1.6 \\
Qwen2.5-Coder-32B & 8.2 & 2.6 & 81.8 & 7.4 & 9.5 & 1.1 \\
Qwen2.5-72B-Instruct & 12.1 & 3.6 & 80.7 & 3.7 & 13.8 & 1.4 \\
\textbf{WebGen-LM-7B} & 24.9 & 7.1 & 68.0 & 0.0 & 28.4 & 2.5 \\
\textbf{WebGen-LM-14B} & 25.0 & 8.7 & 66.3 & 0.0 & 29.4 & 2.5 \\
\textbf{WebGen-LM-32B} & 34.2 & 8.0 & 57.8 & 0.0 & \textbf{38.2} & 2.8 \\
\midrule
\textbf{OpenHands} \\
\midrule
Claude-3.5-Sonnet    & 18.1 &  8.3 & 58.6 & 15.0 & 22.3 & 2.6 \\
DeepSeek-R1          &  8.5 &  3.4 & 60.4 & 27.7 & 10.2 & 1.4 \\
Deepseek-V3 & 7.4 & 3.2 & 73.9 & 15.5 & 9.0 & 1.5 \\
\midrule
\textbf{Aider} \\
\midrule
Claude-3.5-Sonnet    & 19.9 &  5.9 & 42.0 & 32.1 & 22.9 & 1.9 \\
DeepSeek-R1          & 23.3 &  8.7 & 44.5 & 23.5 & 27.7 & 2.7 \\
Deepseek-V3 & 12.5 & 3.1 & 54.3 & 30.1 & 14.1 & 1.2 \\
\bottomrule
\end{tabularx}
\label{tab:main_results}
\end{table}

\begin{table}[t]\fontsize{9}{10}\selectfont
\centering
\caption{Category-wise evaluation results. The first three columns represent categories of website-generation instructions, while the last three represent categories of test cases. The highest score in each category is marked in bold.}
\begin{tabularx}{\textwidth}{>{\raggedright\arraybackslash\hsize=1.4\hsize}X
>{\centering\arraybackslash\hsize=0.6\hsize}X
>{\centering\arraybackslash\hsize=0.6\hsize}X
>{\centering\arraybackslash\hsize=0.6\hsize}X
>{\centering\arraybackslash\hsize=0.6\hsize}X
>{\centering\arraybackslash\hsize=0.6\hsize}X
>{\centering\arraybackslash\hsize=0.8\hsize}X}
\toprule
\multirow{2}{\hsize}{\textbf{Test Name}} & \multicolumn{3}{c}{\textbf{Instruction Categories}} & \multicolumn{3}{c}{\textbf{Test Case Categories}} \\
\cmidrule(r){2-4} \cmidrule(r){5-7} 
 & \textbf{Content Presentation} & \textbf{User Interaction} & \textbf{Data Management} & \textbf{Functional Testing} & \textbf{Data Display Testing} & \textbf{Design Validation Testing} \\
\midrule
\textbf{Bolt.diy} \\
\midrule
Claude-3.5-Sonnet & 35.6 & 21.2 & 26.2 & 17.1 & 26.3 & 52.0 \\
DeepSeek-R1 & 43.7 & 20.6 & 24.7 & 21.1 & 29.3 & 44.3 \\
DeepSeek-V3 & 37.1 & 16.6 & 11.2 & 10.5 & 28.2 & 38.1 \\
GPT-4o & 26.4 & 5.9 & 11.2 & 4.7 & 19.6 & 24.6 \\
o3-mini & 28.7 & 17.7 & 13.4 & 11.4 & 25.5 & 33.6 \\
Qwen2.5-Coder-32B & 17.5 & 6.9 & 5.9 & 1.9 & 14.5 & 23.0 \\
Qwen2.5-72B-Instruct & 28.2 & 10.1 & 5.6 & 5.8 & 21.0 & 25.4 \\
\textbf{WebGen-LM-7B} & 27.9 & 23.8 & 38.1 & 22.0 & 27.7 & 47.5 \\
\textbf{WebGen-LM-14B} & 30.2 & 27.8 & 31.6 & 23.6 & 26.9 & 49.2 \\
\textbf{WebGen-LM-32B} & \textbf{46.6} & \textbf{33.2} & \textbf{38.8} & \textbf{29.1} & \textbf{43.0} & \textbf{56.1} \\
\midrule
\textbf{OpenHands} \\
\midrule
Claude-3.5-Sonnet & 32.8 & 18.4 & 18.4 & 12.4 & 33.9 & 32.0 \\
DeepSeek-R1       & 16.4 &  8.9 &  5.9 &  5.0 &  9.9 & 25.0 \\
Deepseek-V3 & 12.6 & 7.3 & 8.4 & 3.8 & 8.1 & 25.0 \\
\midrule
\textbf{Aider} \\
\midrule
Claude-3.5-Sonnet & 31.9 & 21.1 & 16.6 & 14.9 & 30.1 & 34.0 \\
DeepSeek-R1       & 39.1 & 28.6 & 13.4 & 17.6 & 35.2 & 44.3 \\
Deepseek-V3 & 17.8 & 12.8 & 12.5 & 9.7 & 19.1 & 18.4 \\
\bottomrule
\end{tabularx}
\label{tab:categorical_results}
\end{table}

We present the results on the entire WebGen-Bench dataset in Tab.\ref{tab:main_results}, and the accuracy for each category of instructions and test cases in Tab.~\ref{tab:categorical_results}. Accuracy is computed using the formula $\text{Accuracy} = \frac{N_\text{Yes} + 0.5\times N_\text{Partial}}{N_\text{Total}} \times 100\%$, where $N_\text{Yes}$ and $N_\text{Partial}$ denote the number of test cases assessed as YES and PARTIAL, respectively, and $N_\text{Total}$ is the total number of test cases.

\textbf{Main Results.} Based on the experimental results, we make the following observations:
(1) As shown in Tab.~\ref{tab:categorical_results}, WebGen-LM-32B achieves the highest accuracy of 38.2\%, surpassing the best proprietary model, DeepSeek-R1, by 10.4\%, demonstrating the effectiveness of our training set and the rejection-sampling process.
(2) Bolt.diy with DeepSeek-R1 as the engine achieves the highest accuracy among general LLMs at 27.8\%, closely followed by Claude-3.5-Sonnet with an accuracy of 26.4\%. This indicates that the best-performing models are still far from saturating WebGen-Bench, highlighting that our benchmark remains challenging for current LLMs and agent frameworks.
(3) Smaller general open-source models, such as Qwen2.5-Coder-32B and Qwen2.5-72B-Instruct, show a significant performance gap compared to proprietary models.
(4) In terms of appearance scores, Bolt.diy with Claude-3.5-Sonnet achieves the best performance of 3.0. The appearance score exhibits a loose correlation with accuracy, as functional webpages typically do not suffer from major rendering issues.

\textbf{Categorical Results.} Apart from the three main instruction categories (shown in Tab.~\ref{tab:bench_category_number}), we also classify the test cases into three primary categories based on what they are intended to evaluate: Functional Testing, Data Display Testing, and Design Validation Testing. Detailed definitions and statistics for these categories are provided in Fig.~\ref{fig:task_case_main} and Tab.~\ref{tab:task_cases_category_number} in Appendix\ref{sec:test_cases_categories}. As shown in Tab.~\ref{tab:categorical_results}, among the different categories of test cases, Design Validation Testing achieves the highest accuracy in most cases, while Functional Testing generally yields lower accuracy. Among instruction categories, Content Presentation consistently demonstrates the highest accuracies. This indicates that superficial aspects, such as color themes, are easier to implement than deeper internal functionalities.

\subsection{Ablation Studies}

\begin{table}[t]\fontsize{9}{9}\selectfont
\centering
\caption{Alignment between the UI agent testing results and human testing results. The alignment rate denotes the proportion of test cases in which the UI agent’s results match those of human testers.}
\begin{tabularx}{\textwidth}{
  >{\raggedright\arraybackslash\hsize=1.8\hsize}X
  >{\raggedright\arraybackslash\hsize=0.8\hsize}X
  >{\centering\arraybackslash\hsize=0.8\hsize}X
  >{\centering\arraybackslash\hsize=0.8\hsize}X
  >{\centering\arraybackslash\hsize=0.8\hsize}X
  >{\centering\arraybackslash\hsize=0.6\hsize}X
  >{\centering\arraybackslash\hsize=0.8\hsize}X
}
\toprule
\textbf{Model} & \textbf{Testing Method} & \textbf{Yes Rate} & \textbf{Partial Rate} & \textbf{No Rate} & \textbf{Accuracy} & \textbf{Alignment Rate} \\
\midrule
\multirow{2}{\hsize}{Claude-3.5-Sonnet} 
    & UI Agent         & 22.6 & 7.6 & 64.1 & 26.4 & 90.3 \\
    & Manual & 22.4 & 7.1 & 59.0 & 26.0 & --   \\
\midrule
\multirow{2}{\hsize}{Deepseek-R1} 
    & UI Agent         & 24.7 & 6.2 & 64.3 & 27.8 & 86.1 \\
    & Manual & 28.0 & 4.3 & 58.1 & 30.1 & --   \\
\midrule
\multirow{2}{\hsize}{Deepseek-V3} 
    & UI Agent         & 18.5 & 4.5 & 73.9 & 20.8 & 94.4 \\
    & Manual & 19.0 & 4.5 & 70.3 & 21.3 & --   \\
\bottomrule
\end{tabularx}
\label{tab:manual_testing_results}
\end{table}

\begin{figure}[t]
  \centering
  \begin{minipage}[c]{0.55\textwidth}
    \centering
    \captionof{table}{Comparison of yes rate and accuracy at different sample sizes. The base model is Qwen2.5-Coder-32B-Instruct.}
    \begin{tabular}{@{}ccc@{}}
      \toprule
      \textbf{Sample Number} & \textbf{Yes Rate} & \textbf{Accuracy}\\
      \midrule
      150 & 21.8 & 25.1\\
      300 & 28.6 & 31.9\\
      600 & \textbf{34.2} & \textbf{38.2}\\
      \bottomrule
    \end{tabular}
    \label{tab:sample_numbers_accuracy}
  \end{minipage}
  \hfill
  \begin{minipage}[c]{0.40\textwidth}
    \centering
    \includegraphics[width=\linewidth]{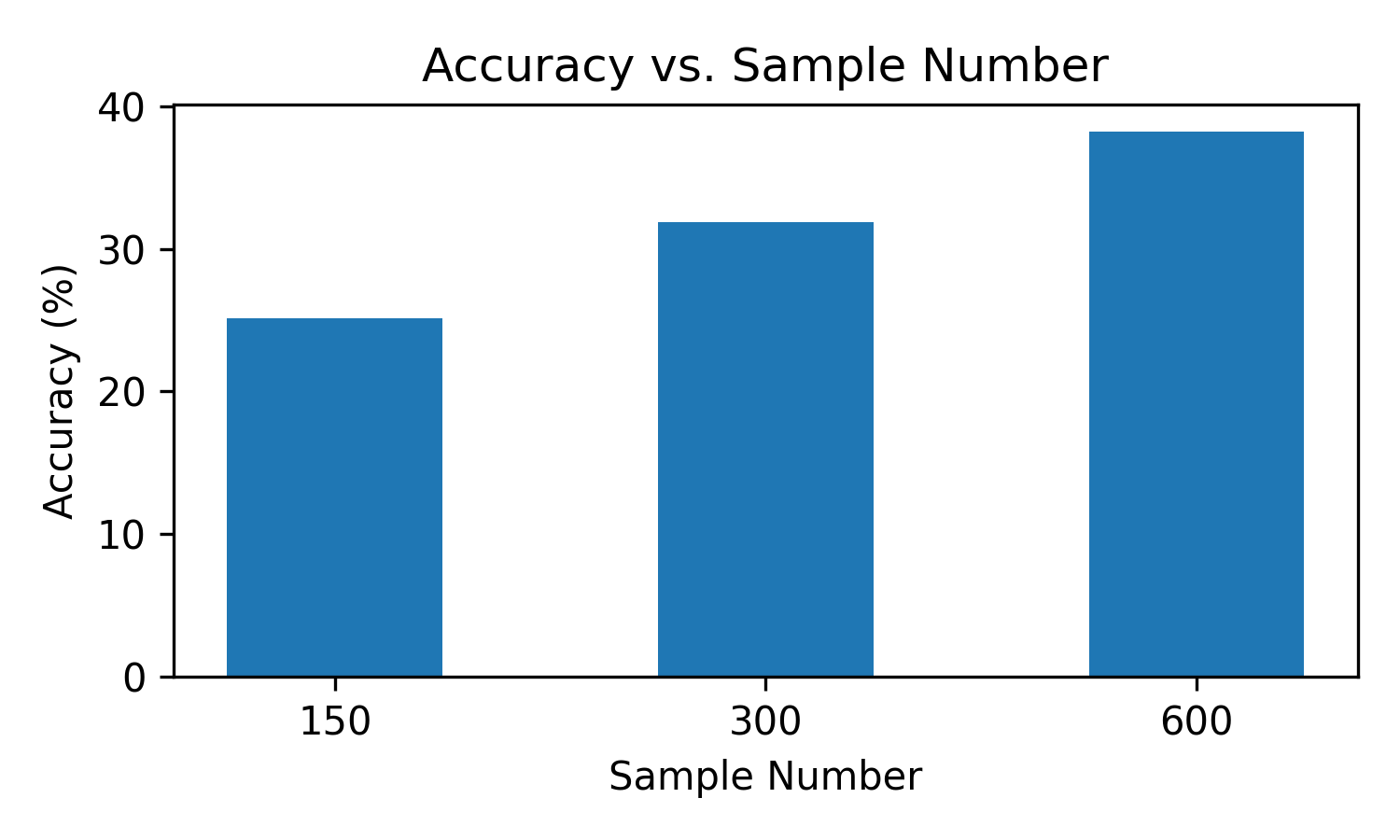}
    \captionof{figure}{Accuracy vs.\ sample number.}
    \label{fig:sample_numbers_accuracy}
  \end{minipage}
\end{figure}

\textbf{Analysis of the Accuracy of UI Agent Testing Results.} To analyze the accuracy of the UI agent testing process, we manually examined three sets of testing results on Bolt.diy. We select the results of Claude-3.5-Sonnet, DeepSeek-R1, and DeepSeek-V3 as the accuracies of these three models are high and are close to each other. The manual testing results serve as the ground truth and require precision; therefore, three human testers independently annotated the results and we assessed the consistency of their annotations. If the annotations of a test case are inconsistent, a fourth human tester is tasked with re-examining the test case and the inconsistent annotations to decide on a final annotation. We present the results of manual testing in Tab.~\ref{tab:manual_testing_results}. The Alignment Rate is computed with $\text{Alignment Rate} = \frac{N_{\text{Manual}=\text{Agent}}}{N_\text{total}} \times 100\%$, where $N_{\text{Manual}=\text{Agent}}$ denotes the number of test cases where the agent-generated result aligns with the manually-annotated result. Further analysis regarding the accuracy of appearance scores is presented in Appendix~\ref{sec:appearance_score_reliability}.

\textbf{Analysis of the Number of Training Samples.} We analyze the effect of the number of training samples on the accuracy of the fine-tuned models. Specifically, we fine-tune Qwen2.5-Coder-32B-Instruct using 150, 300, and 600 samples, respectively. As shown in Fig.~\ref{fig:sample_numbers_accuracy} and Tab.~\ref{tab:sample_numbers_accuracy}, accuracy consistently increases with the number of training samples, highlighting the potential of our training set. We did not sample additional trajectories due to API budget constraints. Nevertheless, the current sample size already demonstrates the effectiveness of WebGen-Instruct for training website generation LLMs. Further accuracy improvements through additional data or techniques such as data augmentation are left for future work.

\section{Conclusion}

In this paper, we introduce WebGen-Bench, a novel benchmark for evaluating the ability of LLM-based agents to generate websites from scratch. The benchmark requires agents to construct and organize multi-file codebases while satisfying various functional and visual constraints. We evaluate three code-agent frameworks using both proprietary and open-source LLMs. The best-performing combination, Bolt.diy with DeepSeek-R1, achieves an accuracy of only 27.8\%, highlighting the challenging nature of our benchmark. Additionally, we construct a training set of 6,667 website-generation instructions and fine-tune Qwen2.5-Coder-32B on 600 Bolt.diy trajectories generated by DeepSeek-V3, resulting in an accuracy of 38.2\%—surpassing even the best proprietary model.

\newpage
\bibliographystyle{plain}
\bibliography{ref}

\newpage
\appendix

\section{Ethics Statement}

The WebGen-Bench dataset is entirely composed of synthetically generated instructions and test cases, curated manually and synthesized using artificial intelligence. While this resource is non-commercial, we emphasize that its construction process maintains a clear distance from potential ethical or legal concerns, particularly regarding intellectual property.
 
\textbf{Legal compliance.} We take great care in our methodology to uphold copyright integrity, utilizing three protective approaches to safeguard against infringement: (1) all base project descriptions originate from the creative efforts of the authors and student volunteers; (2) the 20 fundamental categories are sufficiently abstracted through systematic analysis; and (3) our framework does not copy content from existing websites or platforms, thereby avoiding copyright infringement risks associated with specific commercial implementations.

\textbf{Dataset Intended Usage and License.} We document the WebGen-Bench dataset in this paper and note that both the dataset and the code used for reproducing results are publicly available. We intend for researchers to use this dataset to better evaluate the ability of LLM-based agents to generate websites from scratch. We take full responsibility in the event of any rights violations. The WebGen-Bench dataset and our open-source code are released under the MIT license.

\section{Limitations and Future Work}

Website generation in this work is primarily conducted using TypeScript, JavaScript, CSS, and HTML. Other languages such as Python, Java, and Go are not used, due to the complexity of integrating them into the agent framework. Expanding the range of supported languages and tools for automatic website generation with code agents is a promising direction for future research. Additionally, we only employed supervised fine-tuning to enhance the performance of open-source LLMs on website generation, without utilizing other post-training strategies such as reinforcement learning or direct preference optimization~\citep{rafailov2023direct}. These methods present valuable opportunities for future exploration.

\section{Prompt for Deriving Instructions from Website Development Project Descriptions}
\label{sec:prompt_derive_instruction}

Fig.~\ref{fig:instruction_generation} presents the prompt used to derive website-generation instructions from web development project descriptions created by human annotators. Notably, the model is instructed to exclude any requirements related to technical implementation details, as the goal is to evaluate the code agents' ability to make such decisions independently.

\begin{figure}
\begin{tcolorbox}[colback=blue!5!white,colframe=blue!75!black]
\begin{small}
\textbf{Prompt:}

<task>

You will be given a piece of text containing the basic information of a web development project. The information involves a main objective and a list of functional and appearance requirements. You are requested to convert the information into instructions to build a web application. You should output a detailed multi-sentence instruction in English explaining in detail the different functions the applications should have.

</task>

\vspace{2mm}

<important>

1. Your output should align with the main objective of the website and expand upon the requirements.

2. You should not specify any technical details in the instructions.

3. You should not refer to any outside applications in your instructions.

4. You should not output any additional comments.

</important>

\vspace{2mm}

The following is an example:

\vspace{2mm}

<example>

\vspace{2mm}

Objective:

A hotel and travel ticket distribution website.

Other requirements:

1. User login

2. Order tickets and hotels

3. Cancel orders

4. Verify orders

5. Browse tickets and hotels

6. Light blue background and dark olive green component

\vspace{2mm}

Converted Instruction:

\vspace{2mm}

Please implement a distribution website for travel and ticketing that sells products such as tickets and hotels. The website should have functionalities for placing, canceling, and verifying orders. Users should be able to log in, browse products like tickets and hotels, place orders for selected products, cancel selected orders, and verify consumption records. Use light blue in the background layer and dark olive green for the component layer.

</example>

\vspace{2mm}

\vspace{2mm}

Objective:

\{Objective\}

Other requirements:

\{Other requirements\}

\vspace{2mm}

Converted Instruction:

\end{small}
\end{tcolorbox}
\caption{The prompt for deriving instructions from human annotated descriptions.}
\label{fig:instruction_generation}
\end{figure}

\section{Details of the Decontamination Process}
\label{sec:decontamination}

In this section, we introduce the methods we used to decontaminate the training set from the testing set. We first employ 5-gram Jaccard similarity, removing the instructions in the training set with a similarity score higher than 0.6 with one of the instructions in the testing set. Then, to remove the instructions that are semantically similar to those in the testing set, we compute the sentence embeddings of the instructions using the all-MiniLM-L6-v2 model of Sentence-Transformer~\citep{reimers-2019-sentence-bert}, and compute the cosine similarity of the embeddings. We experimented with various threshold settings, and finally settled on removing the training instructions with a cosine similarity of larger than 0.55.

We then inspect whether the remaining training samples contain instructions that are semantic duplicates of the instructions in the testing. For each testing instruction, we retrieve the top-3 training instructions with the highest cosine similarity, and manually inspect them for semantic duplication. We found that the retrieved training samples are all completely different from the testing samples, proving that the final training set is not contaminated. The first three samples in WebGen-Bench and their top matches in the training set are shown in Fig.~\ref{fig:test_sample1_and_matches}, Fig.~\ref{fig:test_sample2_and_matches}, and Fig.~\ref{fig:test_sample3_and_matches}, respectively. The matches are completely different from the test samples.

\begin{figure}[ht]
\centering
\footnotesize
\renewcommand{\arraystretch}{1.4}
\begin{tabular}{p{0.95\textwidth}}
\hline
\rowcolor{lightblue}
\textbf{Test Instruction 1:} Please implement a website for generating stock reports to provide stock information and analysis. The website should have the functionality to search and summarize stock information, and generate customized stock reports based on user requirements. Users should be able to input stock codes or names, select report formats and content, and the website will automatically generate the corresponding reports. The reports should include basic stock information, market trends, financial data, and more. Set the background color to white and the component color to navy. \\
\midrule
\rowcolor{lightgray}
\textbf{Match 1:} Please implement a website for generating PDF reports that creates PDF files containing directories, word clouds, logos, and chart displays. The website should have functionalities for uploading data, selecting templates, customizing content, previewing, and downloading PDFs. Users should be able to upload relevant data, choose from different templates, customize the report content, preview the generated PDF file, and download the final PDF report. Specify bisque as the base color and dark salmon for all components. \newline
\textbf{Similarity:} 0.549 \\
\rowcolor{white}
\textbf{Match 2:} Please implement an accounting factory website for enterprise financial management and statistics. The website should have functionalities for creating service enterprises, setting declaration types, and extracting statistics by quarter and year. Users should be able to log in, create and manage service enterprises, set declaration types, view and analyze financial data, and perform WeChat payment and other operations. Set page background to light beige; color all components with sienna. \newline
\textbf{Similarity:} 0.542 \\
\rowcolor{lightgray}
\textbf{Match 3:} Please implement a report frontend website to display vehicle inspection report data. The website should have functionalities for displaying report templates, inspection report information, and audit status. Users should be able to log in, browse, and view inspection reports, including report details, inspection results, and audit status. Use powder blue for container backgrounds and royal blue for component visuals. \newline
\textbf{Similarity:} 0.538 \\
\hline
\end{tabular}
\caption{Top semantic matches for the first test instruction in WebGen-Bench with similarity scores.}
\label{fig:test_sample1_and_matches}
\end{figure}

\begin{figure}[ht]
\centering
\footnotesize
\renewcommand{\arraystretch}{1.4}
\begin{tabular}{p{0.95\textwidth}}
\hline
\rowcolor{lightblue}
\textbf{Test Instruction 2:} Please implement a web-based neighborhood mapping application for comparing data across different areas. The application should allow users to compare demographic, economic, and crime data across different areas. The application should also include data dashboards with interactive charts and customizable layouts. Use ivory for the background and forest green for components. \\
\midrule
\rowcolor{lightgray}
\textbf{Match 1:} Please implement a geographic spatial data processing website for handling and analyzing geographic spatial data. The website should have functionalities for data conversion, file interpolation, data operation, and data extraction. Users should be able to upload geographic spatial data files, choose different data formats for conversion, perform data interpolation and operation, and extract the required data. The website should also provide data visualization functionality, allowing users to view and analyze geographic spatial data. Assign mint frost as the background color and apply seagreen to all elements. \newline
\textbf{Similarity:} 0.546 \\
\rowcolor{white}
\textbf{Match 2:} Please implement a geographic information system website for displaying maps and managing the backend. The website should have map visualization capabilities to display different types of geographic information. The backend management platform should be able to manage users, permissions, roles, menus, and support specific business management, such as setting up construction orders, inspecting and evaluating drainage facilities, and managing facilities. Users should be able to log in, browse maps, manage backend data, and perform related operations. Set all pages to have a cream background and dark orange components. \newline
\textbf{Similarity:} 0.542 \\
\rowcolor{lightgray}
\textbf{Match 3:} Please develop a Boundary Hunter app to provide nearby data research services. The app should have functionalities for data research, report generation, and user management. Users should be able to log in, browse nearby data research projects, submit research requests, and view reports. The app should also have automated testing and stress testing capabilities to ensure its stability and performance. Use ghost white for the outer layout and cadet blue for UI blocks. \newline
\textbf{Similarity:} 0.542 \\
\hline
\end{tabular}
\caption{Top semantic matches for the second test instruction in WebGen-Bench with similarity scores.}
\label{fig:test_sample2_and_matches}
\end{figure}

\begin{figure}[ht]
\centering
\footnotesize
\renewcommand{\arraystretch}{1.4}
\begin{tabular}{p{0.95\textwidth}}
\hline
\rowcolor{lightblue}
\textbf{Test Instruction 3:} Please implement a multi-company dashboard for managing and displaying financial data from multiple companies. The dashboard should be able to collect and display financial information from each company, provide consolidated reports, and support cross-company comparisons and reporting. Users should be able to browse financial data from each company, view consolidated reports, and perform financial management and reporting. Apply mint cream as the background; style all components with teal. \\
\midrule
\rowcolor{lightgray}
\textbf{Match 1:} Please implement a multi-lingual accounting website for managing financial accounts. The website should have functionalities for logging in, registering, recording, querying, and statistical analysis. Users should be able to log in, create, edit, and delete financial accounts, query historical accounts, and analyze financial status. The website should support multiple languages to facilitate use by users of different languages. Configure the background color to spring green, with components using lime green. \newline
\textbf{Similarity:} 0.549 \\
\rowcolor{white}
\textbf{Match 2:} Please implement an enterprise resource planning backend management system for managing internal company data. The system should have user management, permission management, module lists, add, edit, delete, and display functions. Users should be able to log in to the system, browse and manage data in different modules, including adding new data, editing existing data, deleting unnecessary data, and displaying all data. The system should also support Excel import and export functions for convenient batch data operations. Use alabaster as the screen background and dark cyan for component highlights. \newline
\textbf{Similarity:} 0.542 \\
\rowcolor{lightgray}
\textbf{Match 3:} Please implement a data visualization website for a telecommunications company to display company data. The website should have multiple pages, each with different dynamic effects. The website should include various charts and maps, with charts having dynamic refresh effects and maps implementing three-level drill-down functionality. Users should be able to browse different pages and view the company's data and statistical information. Use almond as the screen background and sienna for component highlights. \newline
\textbf{Similarity:} 0.540 \\
\hline
\end{tabular}
\caption{Top semantic matches for the third test instruction in WebGen-Bench with similarity scores.}
\label{fig:test_sample3_and_matches}
\end{figure}

\section{Application Categories of WebGen-Instruct and WebGen-Bench.}
\label{sec:category_number}

Tab.~\ref{tab:category_number} lists the 20 application categories manually summarized by the authors through browsing web development projects on popular platforms that connect programmers with clients seeking custom website solutions, such as Upwork\footnote{https://www.upwork.com}, Freelancer\footnote{https://www.freelancer.com}, and Proginn\footnote{https://www.proginn.com}. These application categories serve as seed ideas for our human annotators during the brainstorming of new application scenarios.

Detailed definition of each category is as follows:

\begin{itemize}
    \item Personal Portfolio Sites: Showcase individual professional projects, achievements, and skills.
    \item Company Brochure Sites: Static or minimally interactive websites providing company information, products, services, and contact details.
    \item Personal Blog Sites: Regularly updated content sites focusing on personal writing, opinions, experiences, and sharing of knowledge.
    \item Social Media Platforms: Applications enabling users to interact socially, share content, and build networks.
    \item Discussion Forums: Platforms facilitating conversations, topic-based discussions, threads, and community interactions.
    \item E-commerce Web Applications: Online platforms designed for buying and selling goods and services, handling transactions, inventory, and payments.
    \item Email Clients: Applications for managing emails, sending, receiving, organizing, and scheduling email communication.
    \item Project Management Tools: Platforms aiding in task organization, scheduling, collaboration, and resource management for projects.
    \item Streaming and Interactive Platforms: Media-centric platforms for video, audio streaming, or interactive media consumption.
    \item CRM Systems: Customer Relationship Management tools designed to manage interactions, sales, customer data, and marketing.
    \item ERP Platforms: Enterprise Resource Planning systems integrating core business processes such as finance, HR, supply chain, and operations.
    \item Internal Tools: Applications focused on internal company operations, communication, and collaboration.
    \item News and Information Sites: Platforms primarily dedicated to delivering news content, articles, and timely updates.
    \item Publishing/Blogging Platforms: Platforms enabling users to publish, edit, and manage content on a large scale.
    \item Analytics Platforms/Dashboards: Applications providing insights through data visualization, including Business Intelligence and Financial Dashboards.
    \item Browser-Based Games: Interactive, entertainment-focused applications running directly in web browsers.
    \item Learning Platforms: Educational platforms providing courses, training materials, quizzes, and learning management systems.
    \item Travel Booking Portals: Platforms allowing users to search, compare, and book travel services like flights, hotels, and car rentals.
    \item Job Search Platforms: Websites connecting job seekers with employers, allowing job postings, applications, and resume management.
    \item Productivity Applications: Tools for productivity tasks like document editing, spreadsheets, presentations, and collaborative work.
\end{itemize}

\begin{table}[t]
\fontsize{9}{11}\selectfont
\centering
\caption{20 application categories manually summarized from popular web-development websites.}
\label{tab:category_number}
\begin{tabularx}{\columnwidth}{>{\raggedright\arraybackslash\hsize=0.5\hsize}X | >{\raggedright\arraybackslash\hsize=0.5\hsize}X } 
\toprule
\textbf{Application Category} & \textbf{Application Category} \\
\midrule
Productivity Applications & Project Management Tools \\
Internal Tools & Company Brochure Sites \\
E-commerce Web Applications & Streaming and Interactive Platforms \\
Analytics Platforms/Dashboards & News and Information Sites \\
Publishing/Blogging Platforms & ERP Platforms \\
Travel Booking Portals & Learning Platforms \\
CRM Systems & Social Media Platforms \\
Discussion Forums & Personal Blog Sites \\
Email Clients & Browser-Based Games \\
Job Search Platforms & Personal Portfolio Sites \\
\bottomrule
\end{tabularx}
\end{table}

\section{Prompt for Creating Website Test Cases}

Fig.~\ref{fig:test_case_construction} presents the prompt used to construct test cases that evaluate whether the generated website fulfills the requirements specified in the corresponding instruction. The prompt emphasizes the importance of ensuring that all functionality and appearance requirements are covered by the generated test cases. Conversely, every test case should directly reflect an aspect of the instruction. This ensures that the website is thoroughly evaluated and that all test cases are valid.

\begin{figure}
\begin{tcolorbox}[colback=blue!5!white,colframe=blue!75!black]
\begin{small}
\textbf{Prompt:}

Act as a testing specialist. Based on the provided prompt below, which was used to generate a website, create a list of 5-10 actionable instructions to test the website's functionality, content accuracy, and user experience. Each instruction must:  

\vspace{2mm}

1. Direct a UI agent to perform a single, atomic task.
2. Include validation criteria.  

3. Align with the goals and features described in the original prompt.  

4. Ensure each task is atomic (tests one function at a time) and avoids combining multiple sub-tasks.  

\vspace{2mm}

Structure each instruction as:  

\vspace{2mm}

Task: Clear, singular task for the UI agent. 

Expected Result: Specific outcome to confirm success.  

\vspace{2mm}

Original prompt:  

\{orig prompt\}

\vspace{2mm}

Focus on testing:  

- Core functionalities (e.g., forms, navigation).

- Content relevance to the prompt's intent.  

- Accessibility and responsiveness. 

- Appearance requirements.

\vspace{2mm}

IMPORTANT: The tasks must directly reflect ALL of the prompt's requirements and ensure each instruction is independent and minimal. You must not include tasks that test functions that are not explicitly required by the original prompt!

\end{small}
\end{tcolorbox}
\caption{The prompt for deriving test cases that covers all the functional and appearance requirements in the instruction. The \{orig prompt\} is replaced with the corresponding website-generation instruction.}
\label{fig:test_case_construction}
\end{figure}

\section{Prompt for Automatic Evaluation of Test Cases Using an UI Agent}
\label{sec:prompt_ui_agent}

Fig.~\ref{fig:agent_starting_prompt} presents the prompt used to instruct the UI agent to perform the operation described in the test case and respond with YES, NO, or PARTIAL, depending on whether the expected outcome is achieved. Fig.~\ref{fig:limit_reached_prompt} shows the prompt used to induce the agent to make a final decision when the maximum number of allowed website interactions has been reached.

\begin{figure}
\begin{tcolorbox}[colback=blue!5!white,colframe=blue!75!black]
\begin{small}
\textbf{Start-Testing Prompt:}

\vspace{2mm}

Task: \{task\}

Expected Result: \{expected result\}

\vspace{2mm}

Instructions:

- Attempt the task as a user would, using the UI elements available.

- Make multiple attempts if needed to try and achieve the expected result.

- Observe whether the expected result is fully, partially, or not at all achieved.

- IMPORTANT: You can at most interact with the website 15 times. If the limit is reached, directly output your answer.

\vspace{2mm}

At the end of your testing, answer only with one of the following:

- YES: if the expected result was fully achieved.

- NO: if the expected result could not be achieved at all.

- PARTIAL: if only some aspects of the expected result were achieved.

\end{small}
\end{tcolorbox}
\caption{The prompt for starting the operation of a test case, where \{task\} is replaced with the operation to be performed, \{expected result\} is replaced with the expected state of the website after the operation is performed.}
\label{fig:limit_reached_prompt}
\end{figure}

\begin{figure}
\begin{tcolorbox}[colback=blue!5!white,colframe=blue!75!black]
\begin{small}
\textbf{Limit-reached Prompt:}

You have reached the maximum number of allowed interactions with the website.

\vspace{2mm}

Please evaluate the outcome of your attempts based on the expected result:

\vspace{2mm}

Expected Result: \{expected result\}

\vspace{2mm}

Now, answer with one of the following:

\vspace{2mm}

- YES: if the expected result was fully achieved during your interactions.

- NO: if the expected result was not achieved at all.

- PARTIAL: if the expected result was only partially achieved.

\vspace{2mm}

Provide your final answer based on your testing experience.

\end{small}
\end{tcolorbox}
\caption{The prompt for inducing an answer when the limit of the number of website interactions is reached, where \{task\} is replaced with the operation to be performed, \{expected result\} is replaced with the expected state of the website after the operation is performed.}
\label{fig:agent_starting_prompt}
\end{figure}

\section{Prompt for Grading Website Appearance}
\label{sec:grade_appearance_prompt}

Fig.~\ref{fig:agent_starting_prompt} shows the prompt used to grade the aesthetics of webpage appearances. The grading vision-language model (GPT-4o in this case) is instructed to consider metrics such as successful rendering, content relevance, layout harmony, and the modernity and visual appeal of the design, and then output a grade ranging from 1 to 5 (the higher, the better).

\begin{figure}
\begin{tcolorbox}[colback=blue!5!white,colframe=blue!75!black]
\begin{small}
\textbf{Appearance-Grading Prompt:}

Instruction:

You are tasked with evaluating the functional design of a webpage that had been constructed based on the following instruction:

\vspace{2mm}

\{instruction\}

\vspace{2mm}

Grade the webpage’s appearance on a scale of 1 to 5 (5 being highest), considering the following criteria:

\vspace{2mm}

  - Successful Rendering: Does the webpage render correctly without visual errors? Are colors, fonts, and components displayed as specified?
  
  - Content Relevance: Does the design align with the website’s purpose and user requirements? Are elements (e.g., search bars, report formats) logically placed and functional?
  
  - Layout Harmony: Is the arrangement of components (text, images, buttons) balanced, intuitive, and clutter-free?
  
  - Modernness \& Beauty: Does the design follow contemporary trends (e.g., minimalism, responsive layouts)? Are colors, typography, and visual hierarchy aesthetically pleasing?

\vspace{2mm}

Grading Scale:

\vspace{2mm}

  - 1 (Poor): Major rendering issues (e.g., broken layouts, incorrect colors). Content is irrelevant or missing. Layout is chaotic. Design is outdated or visually unappealing.
  
  - 2 (Below Average): Partial rendering with noticeable errors. Content is partially relevant but poorly organized. Layout lacks consistency. Design is basic or uninspired.
  
  - 3 (Average): Mostly rendered correctly with minor flaws. Content is relevant but lacks polish. Layout is functional but unremarkable. Design is clean but lacks modern flair.
  
  - 4 (Good): Rendered well with no major errors. Content is relevant and logically organized. Layout is harmonious and user-friendly. Design is modern and visually appealing.
  
  - 5 (Excellent): Flawless rendering. Content is highly relevant, intuitive, and tailored to user needs. Layout is polished, responsive, and innovative. Design is cutting-edge, beautiful, and memorable.

\vspace{2mm}

Task:

Review the provided screenshot(s) of the webpage. Provide a detailed analysis and then assign a grade (1–5) based on your analysis. Highlight strengths, weaknesses, and how well the design adheres to the specifications.

\vspace{2mm}

Your Response Format:

\vspace{2mm}

Analysis: [2–4 paragraphs addressing all criteria, referencing the instruction]

\vspace{2mm}

Grade: [1–5]

\vspace{2mm}

Your Response:
\end{small}
\end{tcolorbox}
\caption{The prompt for grading the appearance of the webpage.}
\label{fig:appearance_grading_prompt}
\end{figure}

\section{Prompt for Adapting OpenHands Paired with CodeActAgent for WebGen-Bench Evaluation}
\label{sec:openhands_prompt}

Figure~\ref{fig:openhands_prompt} presents the prompt used to evaluate OpenHands in combination with CodeActAgent on the WebGen-Bench benchmark.

\begin{figure}
\begin{tcolorbox}[colback=blue!5!white,colframe=blue!75!black]
\begin{small}
\textbf{OpenHands Prompt:}

Create a website app using typescript, html, and css. Your codebase should be able to be setup using 'npm install', and the service should be able to be started using 'npm run dev'.

\vspace{2mm}

\{instruction\}

\end{small}
\end{tcolorbox}
\caption{The prompt for testing OpenHands paired with CodeActAgent on WebGen-Bench.}
\label{fig:openhands_prompt}
\end{figure}

\section{Prompt for Aider to Generate Websites for WebGen-Bench Evaluation}
\label{sec:aider_prompt}

Fig.~\ref{fig:aider_prompt} shows the prompt used by Aider to generate websites for the WebGen-Bench evaluation.

\begin{figure}
\begin{tcolorbox}[colback=blue!5!white,colframe=blue!75!black]
\begin{small}
\textbf{Aider Prompt:}

You are Aider, an expert AI assistant and exceptional senior software developer with vast knowledge across multiple programming languages, frameworks, and best practices.

<system\_constraints>

- You MUST generate the code and files Directly without telling me the implementation plan, just generate the codes and files. 

- No C/C++ compiler, native binaries, or Git

- Prefer Node.js scripts over shell scripts

- Use Vite for web servers and Node.js for backend

- Databases: prefer libsql, sqlite, or non-native solutions

- When for react dont forget to write vite config and index.html to the project

- You MUST generate a complete package.json file with valid package release version.

</system\_constraints>

\vspace{2mm}

\{instruction\}

\vspace{2mm}

Make sure all the files imported are correctly generated, and a complete package.json file with valid package release version exists. Generate the remaining files if needed.

\end{small}
\end{tcolorbox}
\caption{The prompt for aider websites generation.}
\label{fig:aider_prompt}
\end{figure}

\section{Test Case Categories}
\label{sec:test_cases_categories}

\begin{figure*}[t]
    \centering
    \includegraphics[width=1.0\textwidth]{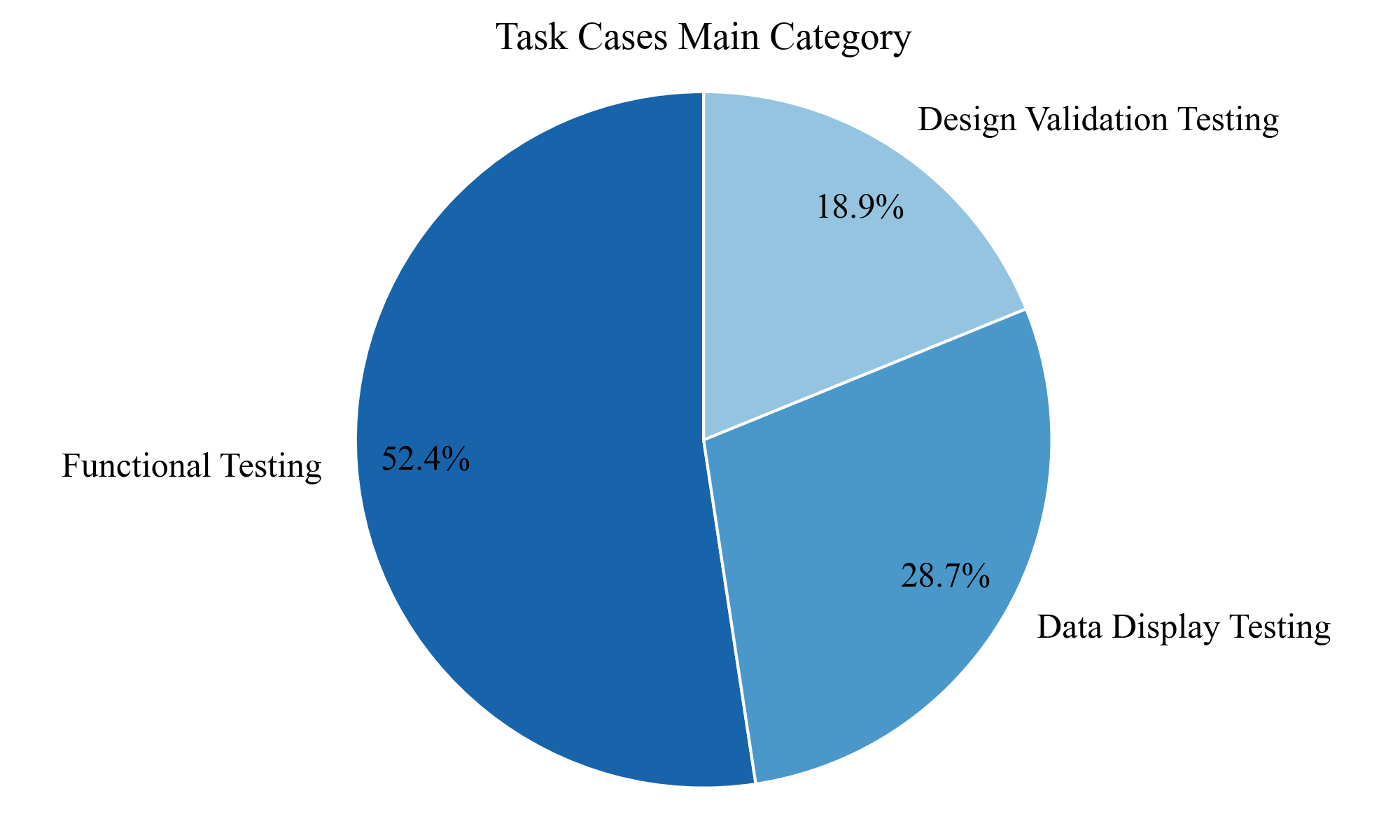}
    \caption{The distribution of the task case categories.}
\label{fig:task_case_main}
\end{figure*}

Fig.~\ref{fig:task_case_main} shows the main category distribution of the task cases. Nearly half of the test cases fall under Functional Testing, around 30\% under Data Display Testing, and approximately 20\% under Design Validation Testing. This is a reasonable distribution, as functional testing typically constitutes the majority of web page evaluations. Additionally, Tab.~\ref{tab:task_cases_category_number} presents the detailed subcategories along with their respective frequencies.

Functional testing ensures that all features of an application work as intended. This includes testing form operations such as submission and validation workflows; verifying authentication flows like user registration, login, and permission checks; and validating payment functionalities in e-commerce checkouts or donation processes. It also encompasses search capabilities across various domains such as stock codes, products, or employees, and filtering data based on specific requirements. Additionally, functional testing covers generation tasks such as creating reports or files; file operations including downloading, uploading, and printing; e-commerce activities such as purchasing or booking items; and communication features like sending messages or emails.

Data display testing focuses on how data is presented and updated within an application. This involves ensuring that dynamic content rendering works correctly, including real-time data updates, website navigation, and page refresh mechanisms. It also includes verifying the accuracy of data visualization elements such as charts, graphs, and maps. Furthermore, this type of testing checks the functionality of displaying detailed information when users request more specific data.

Design validation testing focuses on the aesthetic and responsive aspects of an application's user interface. It involves verifying UI consistency across the application and ensuring that color schemes, typography, and spacing are correctly implemented. Responsive behavior is also tested to confirm that the application adapts properly to different devices and screen sizes. Finally, component styling is checked to ensure that elements such as buttons, icons, and cards adhere to the intended design standards.

\begin{table}[t]
\fontsize{9}{10}\selectfont
\centering
\caption{The number of task cases in each category. There are multiple subcategories under each main category. A task case can belong to one main category and multiple subcategories.}
\label{tab:task_cases_category_number}
\begin{tabularx}{\columnwidth}{>{\raggedright\arraybackslash\hsize=0.6\hsize}X >{\centering\arraybackslash\hsize=0.4\hsize}X | > {\raggedright\arraybackslash\hsize=0.6\hsize}X >{\centering\arraybackslash\hsize=0.4\hsize}X}
\toprule
\textbf{Main Categories}  & \textbf{Task Number} & \textbf{Sub Category} & \textbf{Task Number}  \\
\midrule
\multirow{9}*{Functional Testing} & \multirow{9}*{339} & Form Operations & 134 \\
& & Authentication Flows & 48  \\
& & Payment & 7  \\
& & Searching & 49 \\
& & Filtering & 27  \\
& & Generation & 63  \\
& & File Operation & 23 \\
& & E-commerce & 58  \\
& & Communication & 71 \\
\cmidrule(lr){1-4}
\multirow{3}*{Data Display Testing} & \multirow{3}*{186} & Dynamic Content Rendering & 155 \\
& & Data Visualization & 30  \\
& & Details Information & 91  \\
\cmidrule(lr){1-4}
\multirow{3}*{Design Validation Testing} & \multirow{3}*{122} & UI Consistency & 122 \\
& & Responsive Behavior & 13  \\
& & Component Styling & 9  \\
\cmidrule(lr){1-4}
Total & 667 & & \\
\bottomrule
\end{tabularx}
\end{table}

\section{Analysis of Reliability of Appearance Scores}
\label{sec:appearance_score_reliability}

To strengthen the appearance assessment, we evaluate the bolt.diy results produced by Claude-3.5-Sonnet, DeepSeek-R1, and DeepSeek-V3 with two additional strong multimodal graders—o3 and Claude-3.5-Sonnet. We also manually grade the website screenshots to capture human preference. The results are presented in Tab.~\ref{tab:appearance_score_reliability}. Although o3 and Claude-3.5-Sonnet assign slightly lower scores than GPT-4o, the relative ordering of the three results is unchanged (Claude-3.5-Sonnet > DeepSeek-R1 > DeepSeek-V3). Also, for every set of screenshots, the scores descend in the same order: GPT-4o > o3 > Claude-3.5-Sonnet, showing a consistency in the models’ grading pattern. The ensemble results of the three grading models and the human scores both show the same ranking as GPT-4o. These findings indicate that GPT-4o reliably reflects relative appearance quality and aligns with human preference, making it a suitable choice under a limited budget. We will add this discussion to the revised paper.

\begin{table}[t]\fontsize{9}{10}\selectfont
\centering
\caption{Average appearance scores obtained when the output of each agent–engine LLM is graded by different grading methods.}
\begin{tabularx}{\textwidth}
  {>{\raggedright\arraybackslash\hsize=1.9\hsize}X
   >{\centering\arraybackslash\hsize=0.4\hsize}X
   >{\centering\arraybackslash\hsize=0.3\hsize}X
   >{\centering\arraybackslash\hsize=0.6\hsize}X
   >{\centering\arraybackslash\hsize=0.6\hsize}X
   >{\centering\arraybackslash\hsize=0.4\hsize}X}
\toprule
\textbf{Agent-engine LLM\,$\backslash$\,Grading method} &
\textbf{GPT-4o} & \textbf{o3} & \textbf{Claude-3.5-Sonnet} &
\textbf{Average (ensemble)} & \textbf{Human} \\
\midrule\midrule
Claude-3.5-Sonnet & \textbf{3.0} & 2.7 & 2.4 & 2.7 & 2.8 \\
DeepSeek-R1       & \textbf{2.5} & 2.3 & 2.2 & 2.3 & 2.3 \\
DeepSeek-V3       & \textbf{2.0} & 1.9 & 1.7 & 1.9 & 1.9 \\
\bottomrule
\end{tabularx}
\label{tab:appearance_score_reliability}
\end{table}

\section{Examples of Websites with Different Appearance Scores}
\label{sec:appearance_score_examples}

\begin{figure*}[t]
    \centering
    \includegraphics[width=1.0\textwidth]{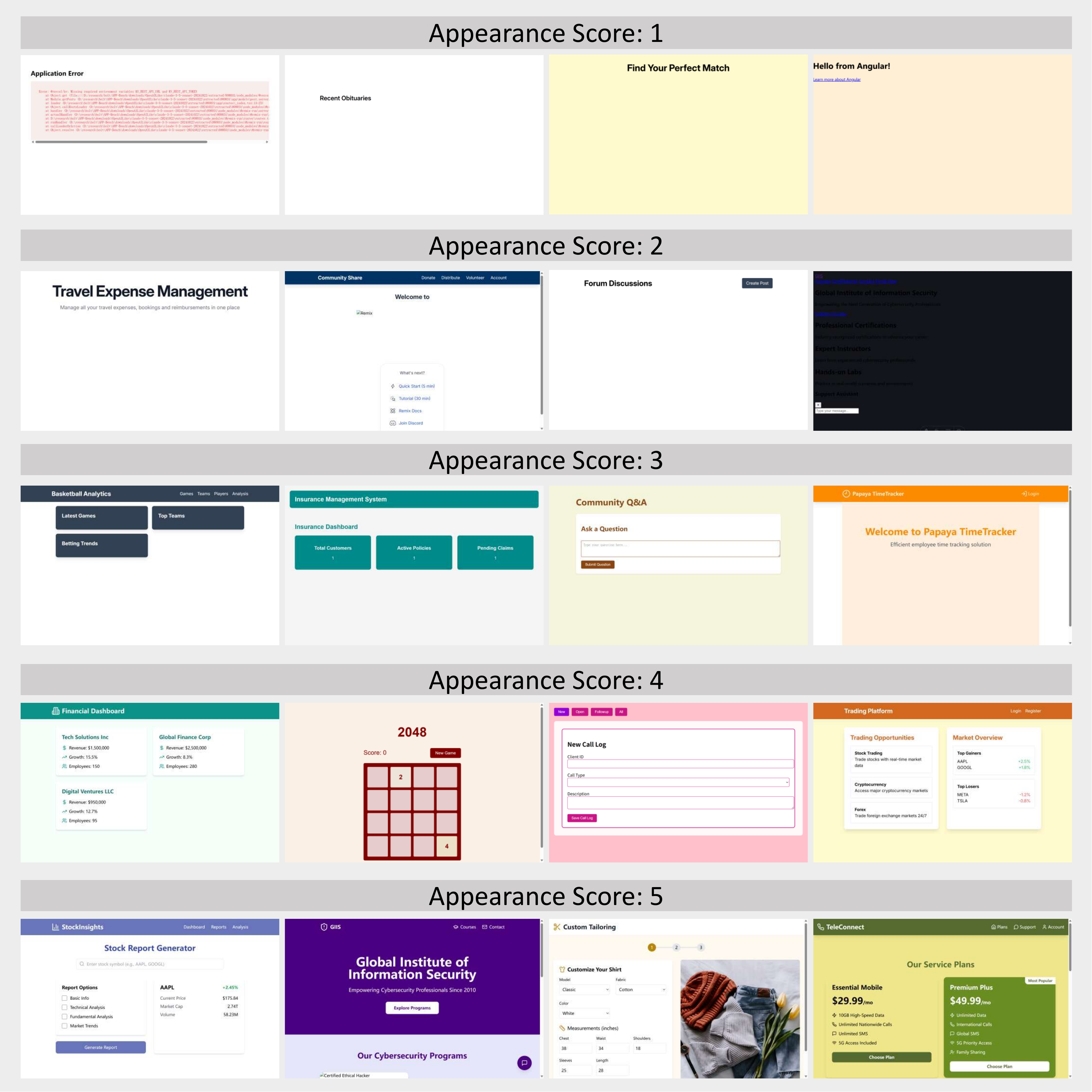}
    \caption{Examples of the screenshots of websites of different appearance scores.}
    
\label{fig:appearance_scores}
\end{figure*}

Fig.~\ref{fig:appearance_scores} presents examples of websites with varying appearance scores. As shown in the figure, the visual quality of the websites improves as the appearance score increases. At a score of one, the websites exhibit major rendering errors or contain irrelevant content, whereas at a score of five, the design appears highly harmonious.

\section{Examples of Websites with Different errors or flaws}
\label{sec:wrong_case_examples}

\begin{figure*}[t]
    \centering
    \includegraphics[width=1.0\textwidth]{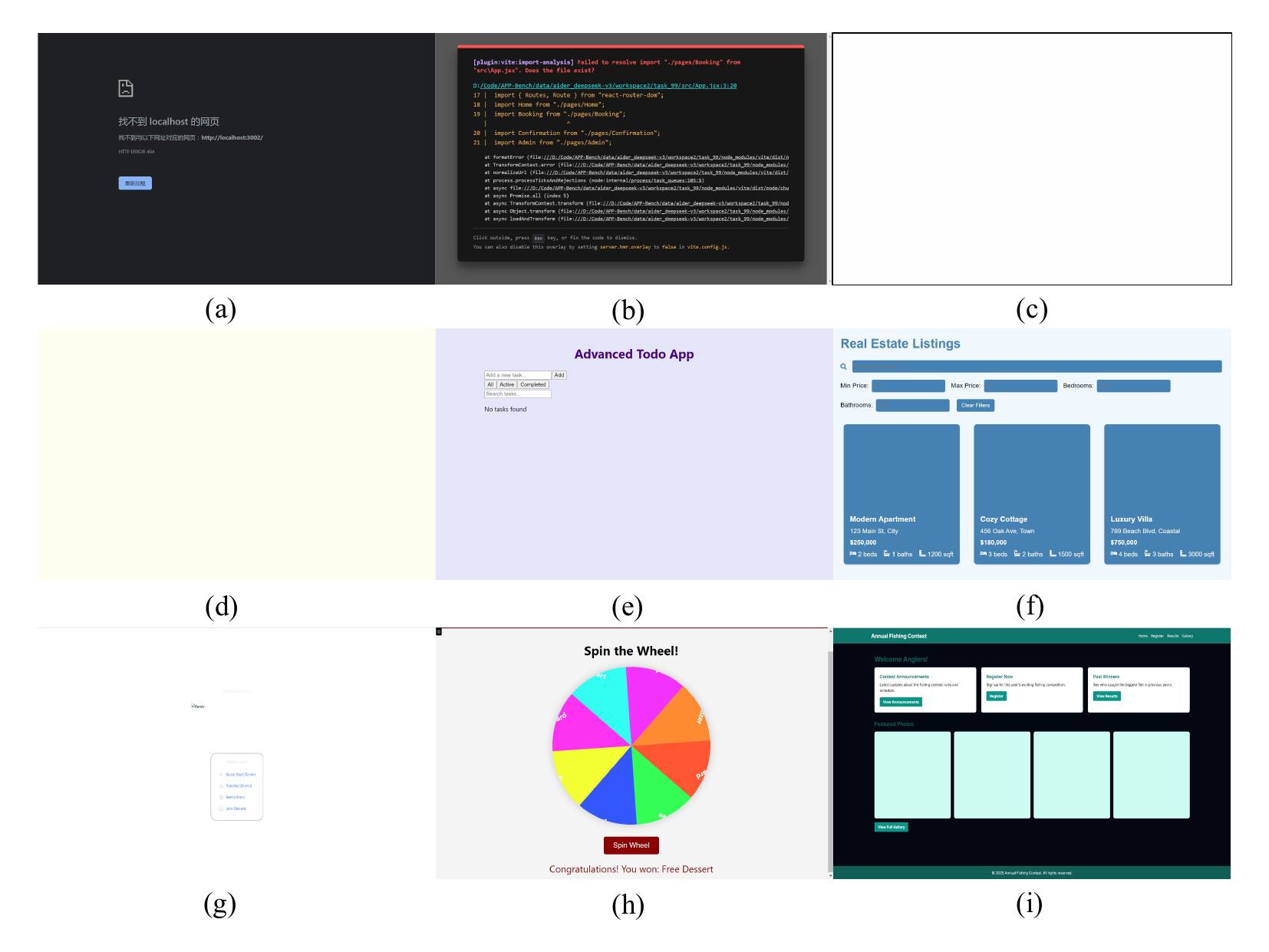}
    \caption{The examples of errors or flaws that generated webs may include.}
    
\label{fig:wrong_case}
\end{figure*}

Fig.~\ref{fig:wrong_case} presents the errors or flaws that a generated website may contain. For example, instances (a), (b), and (c) illustrate three types of errors related to website loading failures. Instances (d), (e), and (f) show incomplete websites: instance (d) displays only the background, instance (e) lacks UI components such as buttons, and instance (f) fails to display an image correctly. Additionally, instance (g) is a website that only uses a template without customization; instance (h) shows incorrect placement of webpage content, such as misaligned text; and instance (i) uses an inappropriate background color.

\section{Examples of UI Agent Testing Processes}

\begin{figure*}[t]
    \centering
    \includegraphics[width=1.0\textwidth]{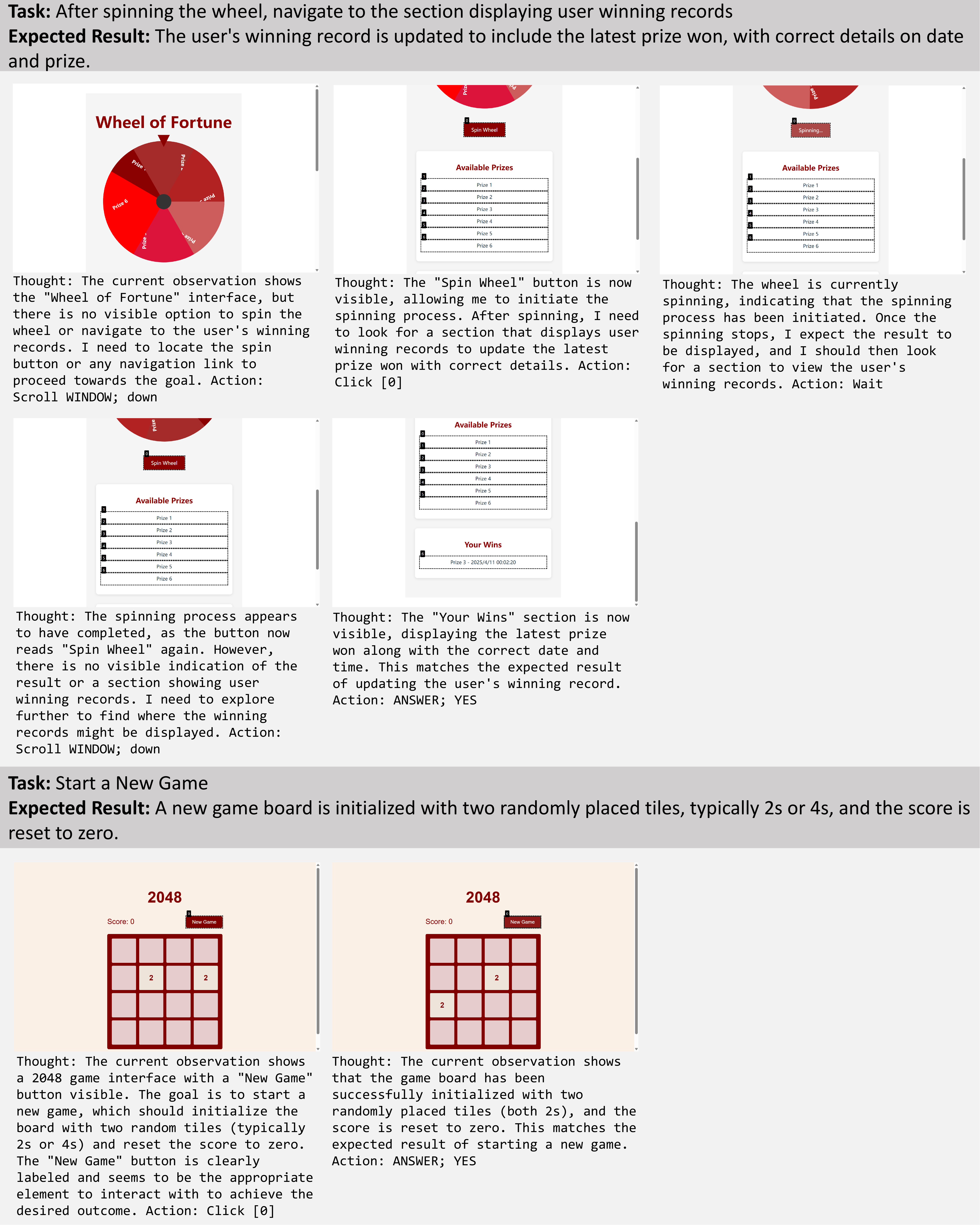}
    \caption{The examples of UI agent testing processes resulting in YES.}
    
\label{fig:yes_examples_1}
\end{figure*}

\begin{figure*}[t]
    \centering
    \includegraphics[width=1.0\textwidth]{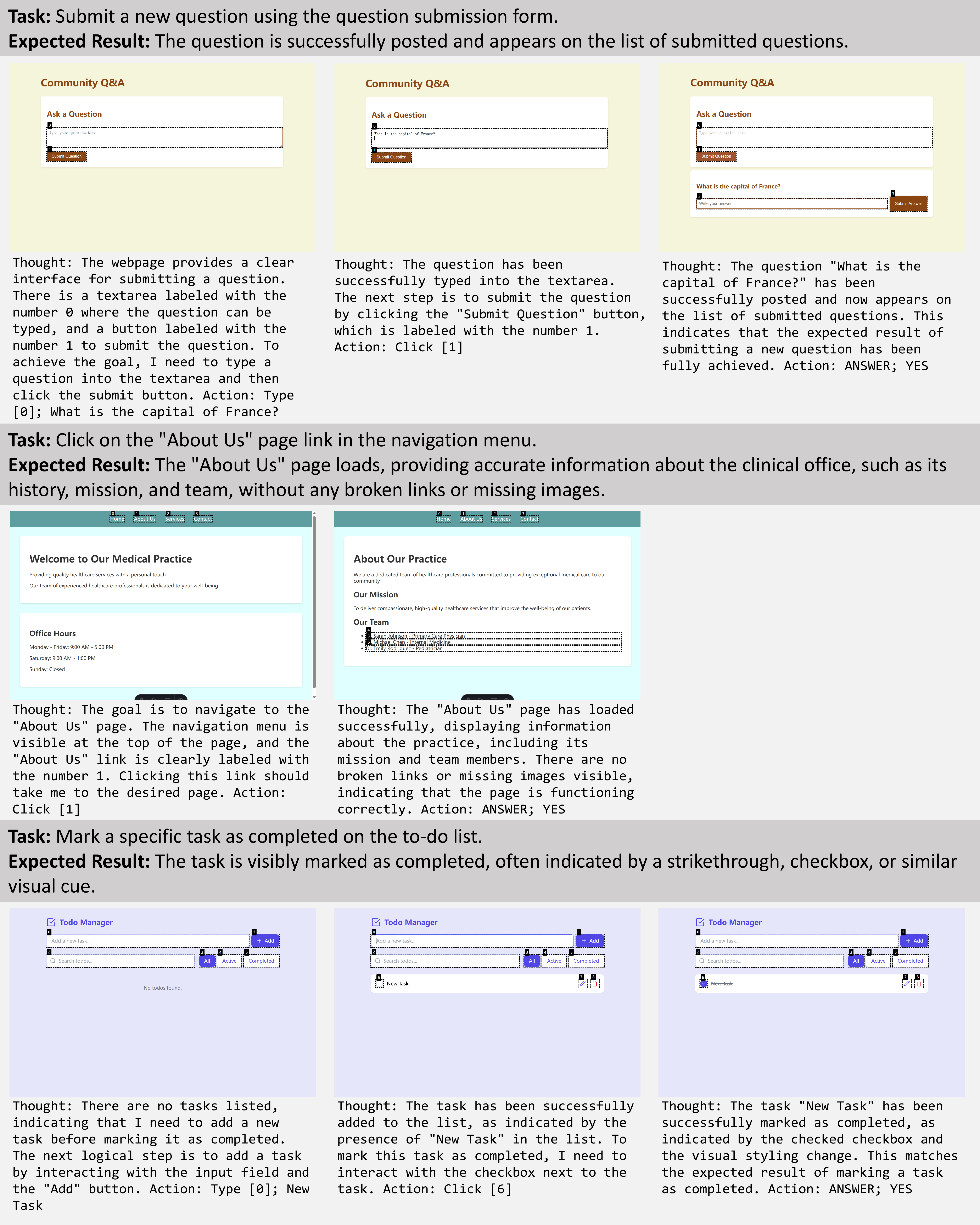}
    \caption{The examples of UI agent testing processes resulting in YES.}
    
\label{fig:yes_examples_2}
\end{figure*}

\begin{figure*}[t]
    \centering
    \includegraphics[width=1.0\textwidth]{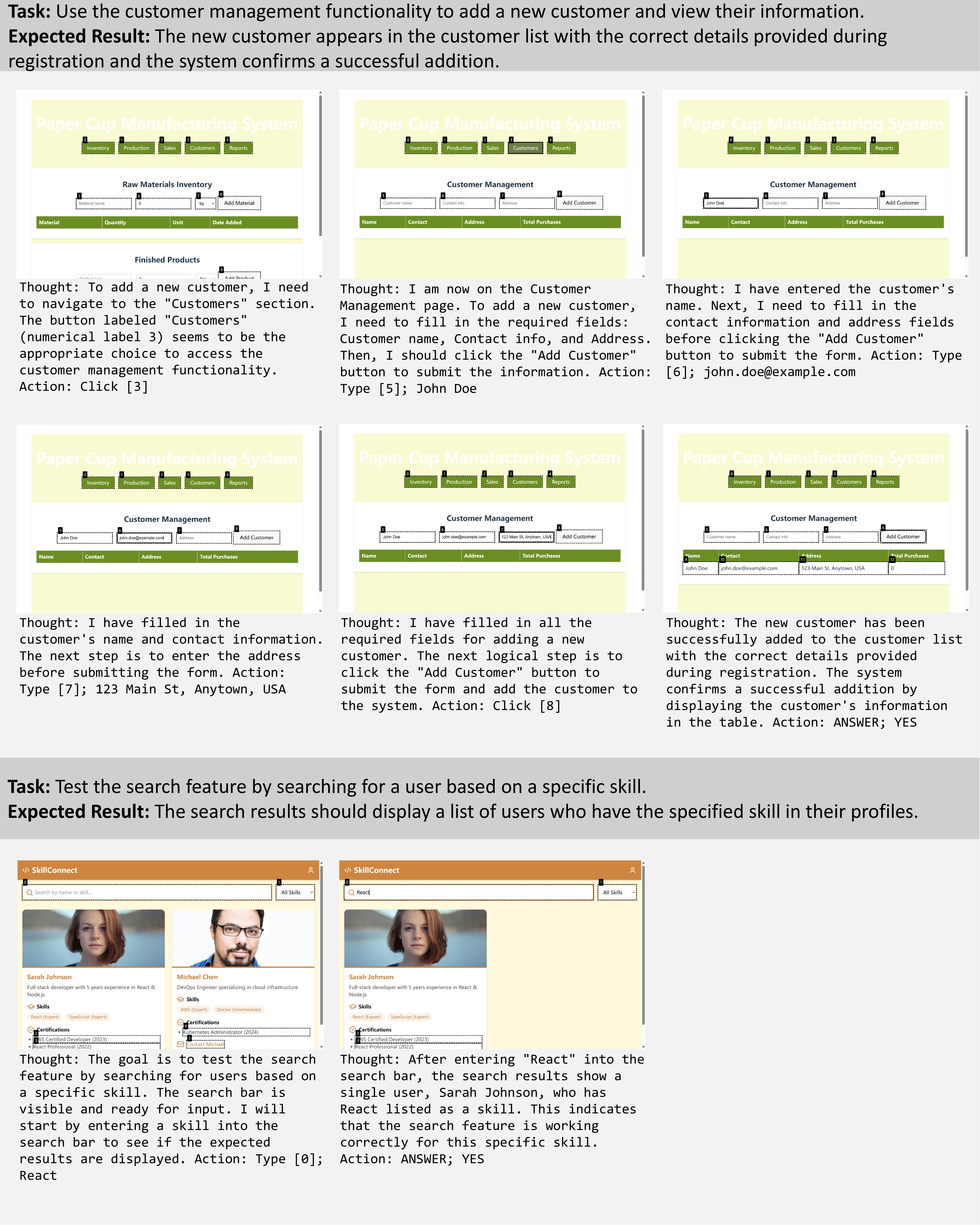}
    \caption{The examples of UI agent testing processes resulting in YES.}
    
\label{fig:yes_examples_3}
\end{figure*}

\begin{figure*}[t]
    \centering
    \includegraphics[width=1.0\textwidth]{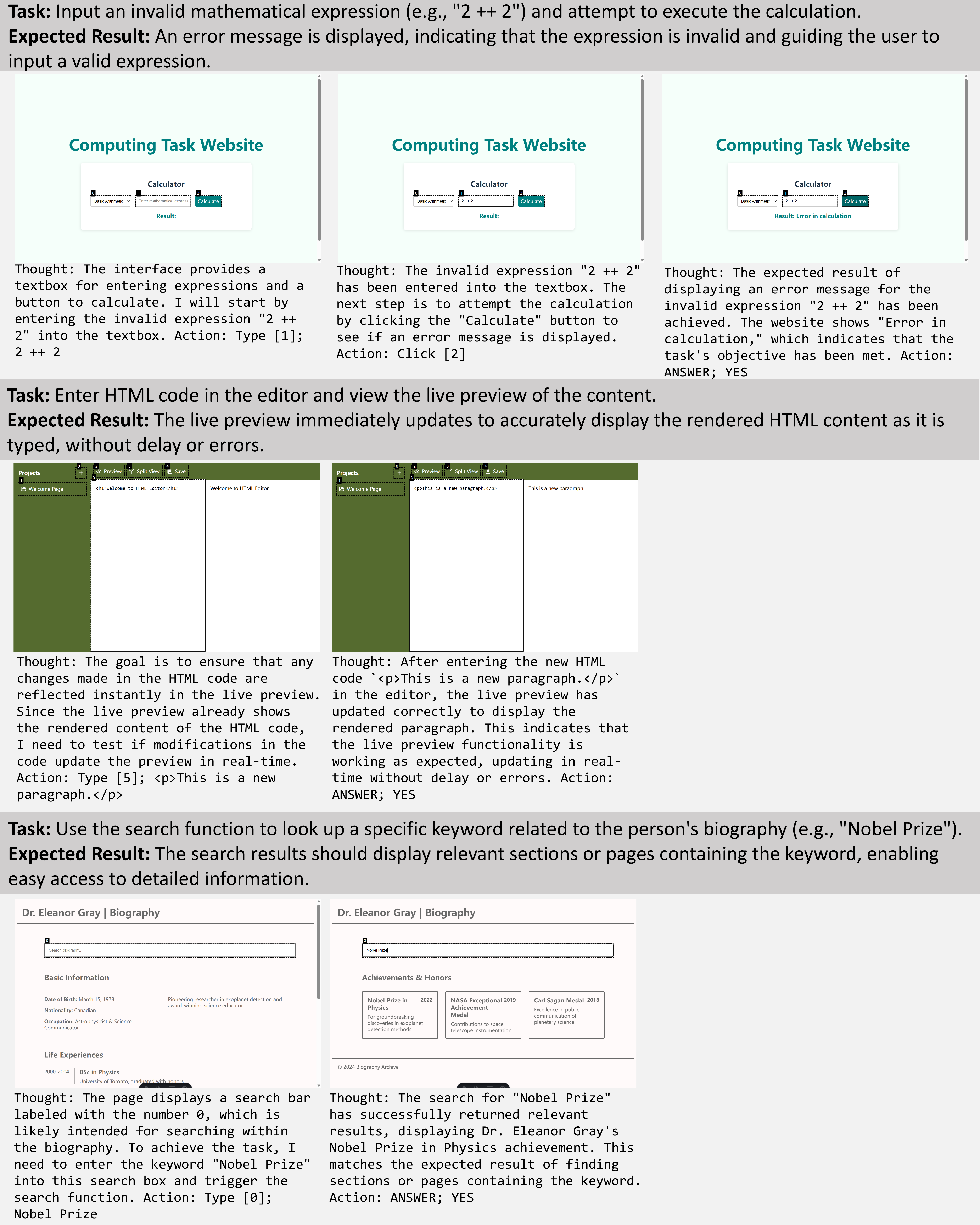}
    \caption{The examples of UI agent testing processes resulting in YES.}
    
\label{fig:yes_examples_4}
\end{figure*}

\begin{figure*}[t]
    \centering
    \includegraphics[width=1.0\textwidth]{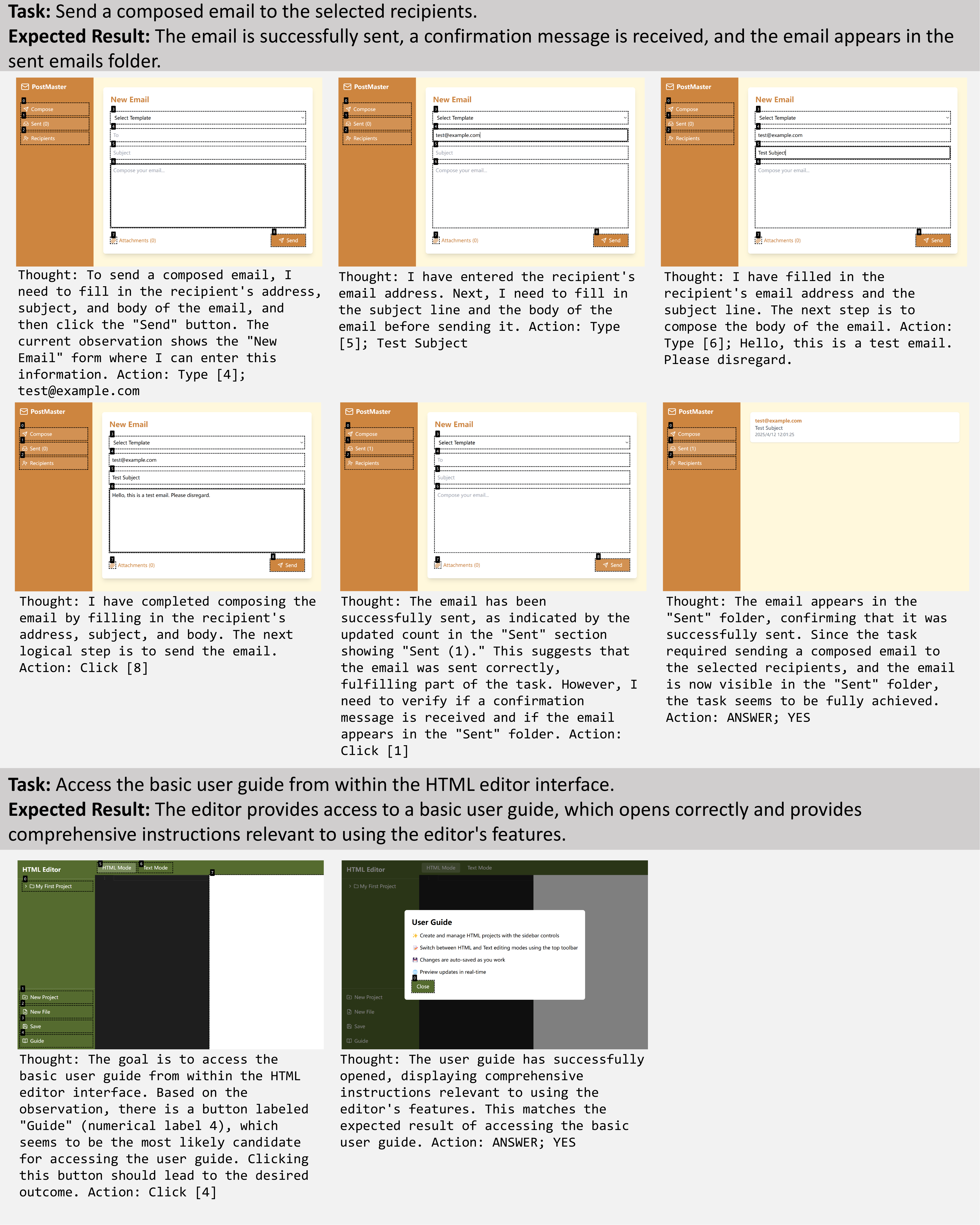}
    \caption{The examples of UI agent testing processes resulting in YES.}
    
\label{fig:yes_examples_5}
\end{figure*}

\begin{figure*}[t]
    \centering
    \includegraphics[width=1.0\textwidth]{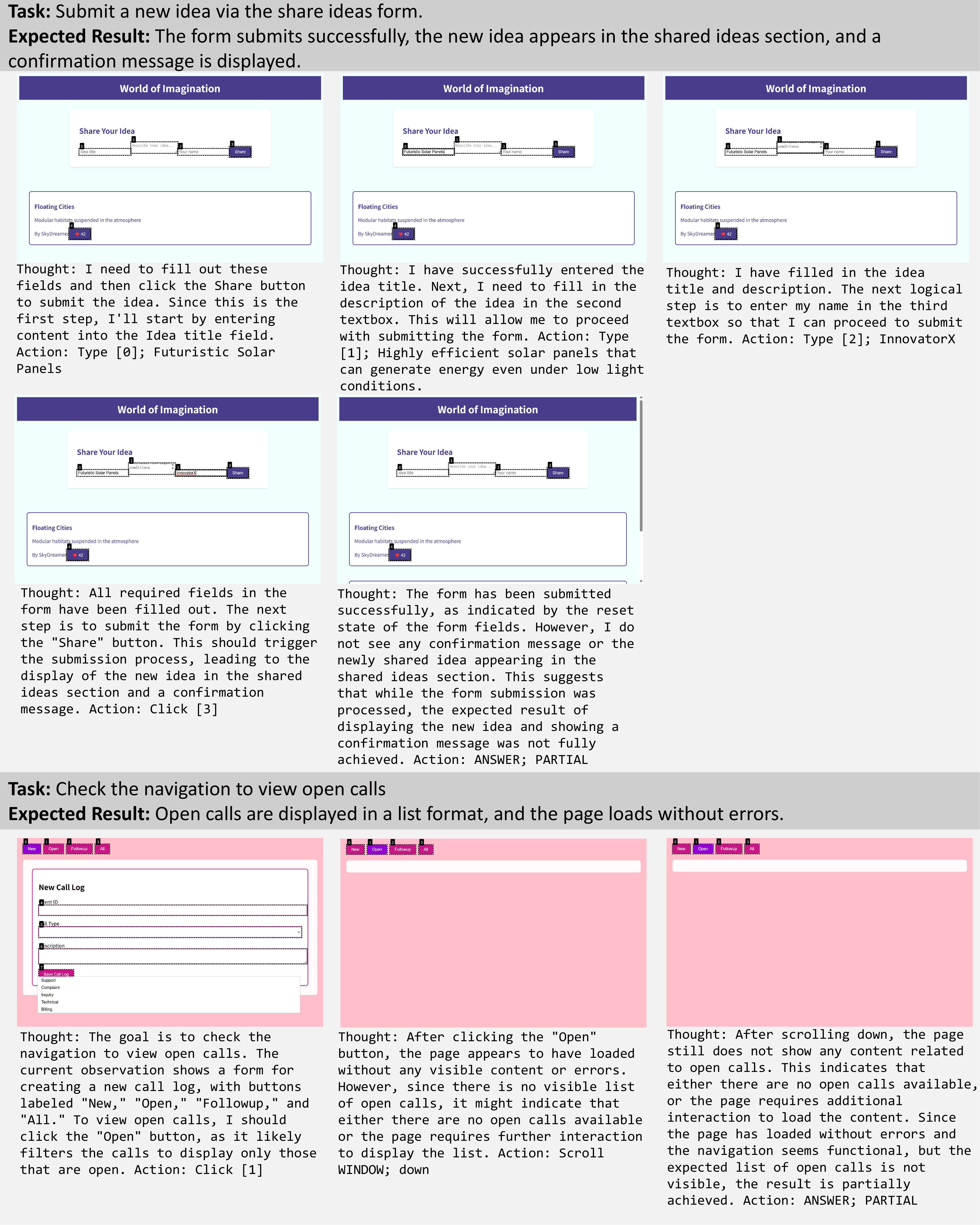}
    \caption{The examples of UI agent testing processes resulting in PARTIAL.}
    
\label{fig:partial_examples_1}
\end{figure*}

\begin{figure*}[t]
    \centering
    \includegraphics[width=1.0\textwidth]{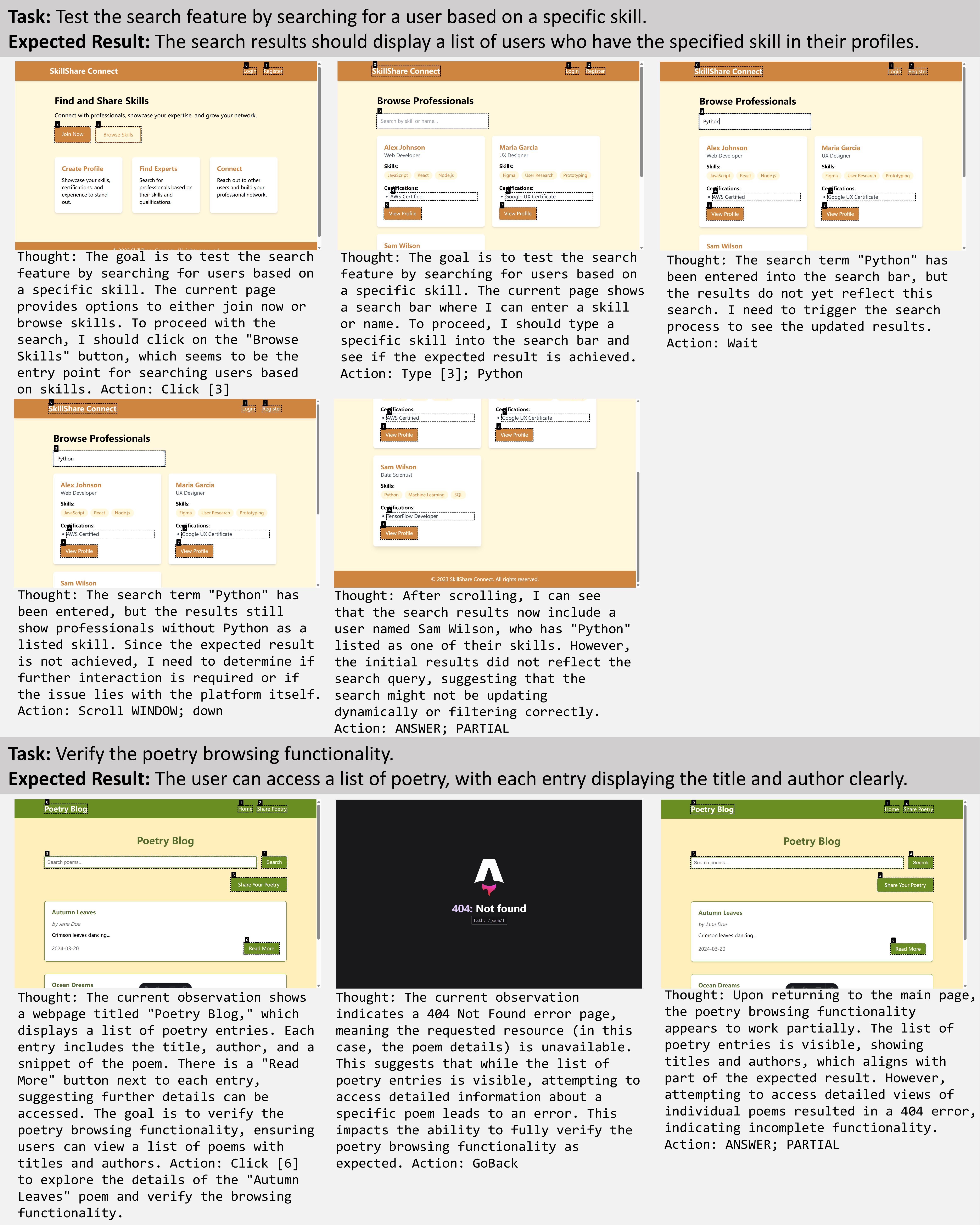}
    \caption{The examples of UI agent testing processes resulting in PARTIAL.}
    
\label{fig:partial_examples_2}
\end{figure*}

\begin{figure*}[t]
    \centering
    \includegraphics[width=1.0\textwidth]{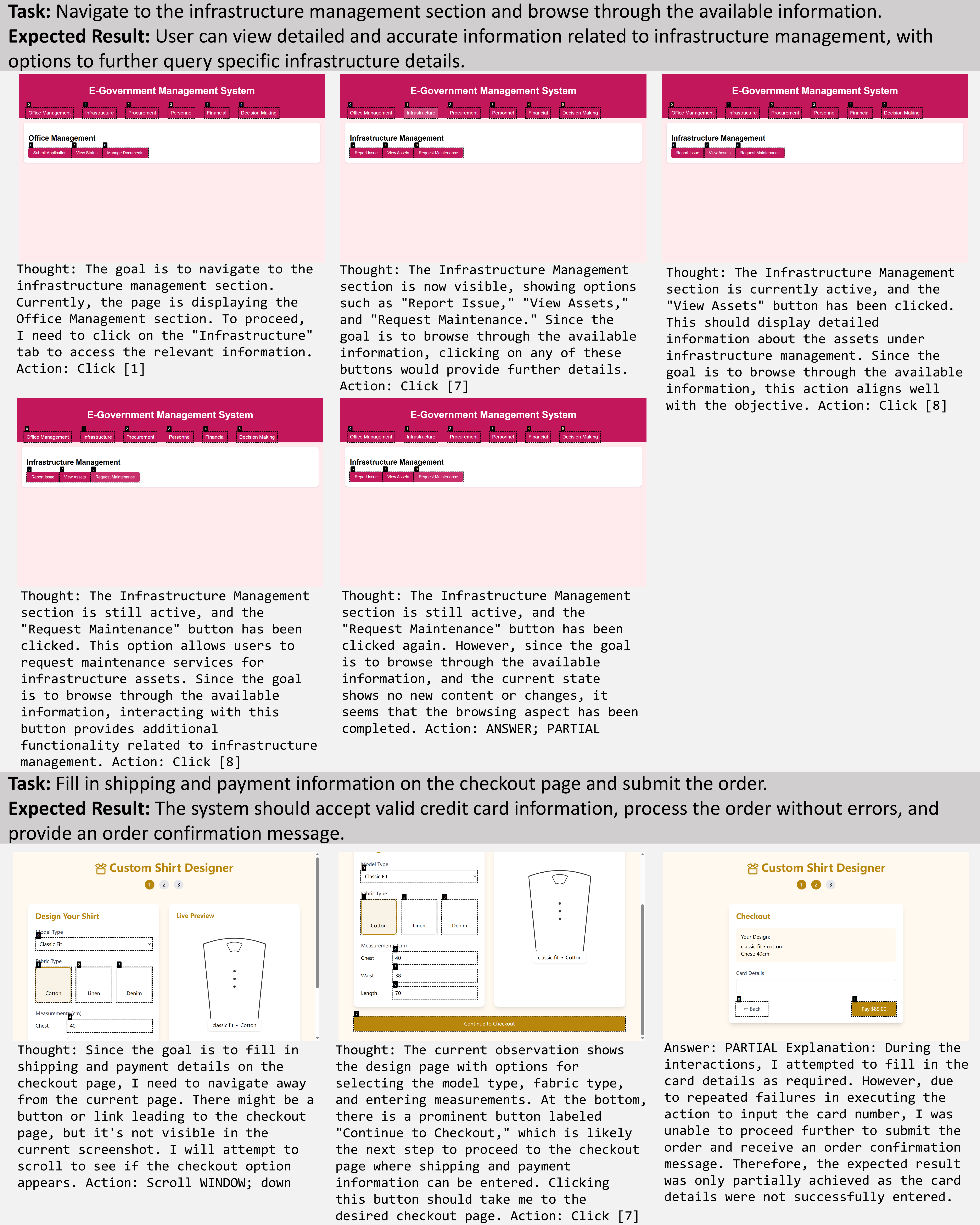}
    \caption{The examples of UI agent testing processes resulting in PARTIAL.}
    
\label{fig:partial_examples_3}
\end{figure*}

\begin{figure*}[t]
    \centering
    \includegraphics[width=1.0\textwidth]{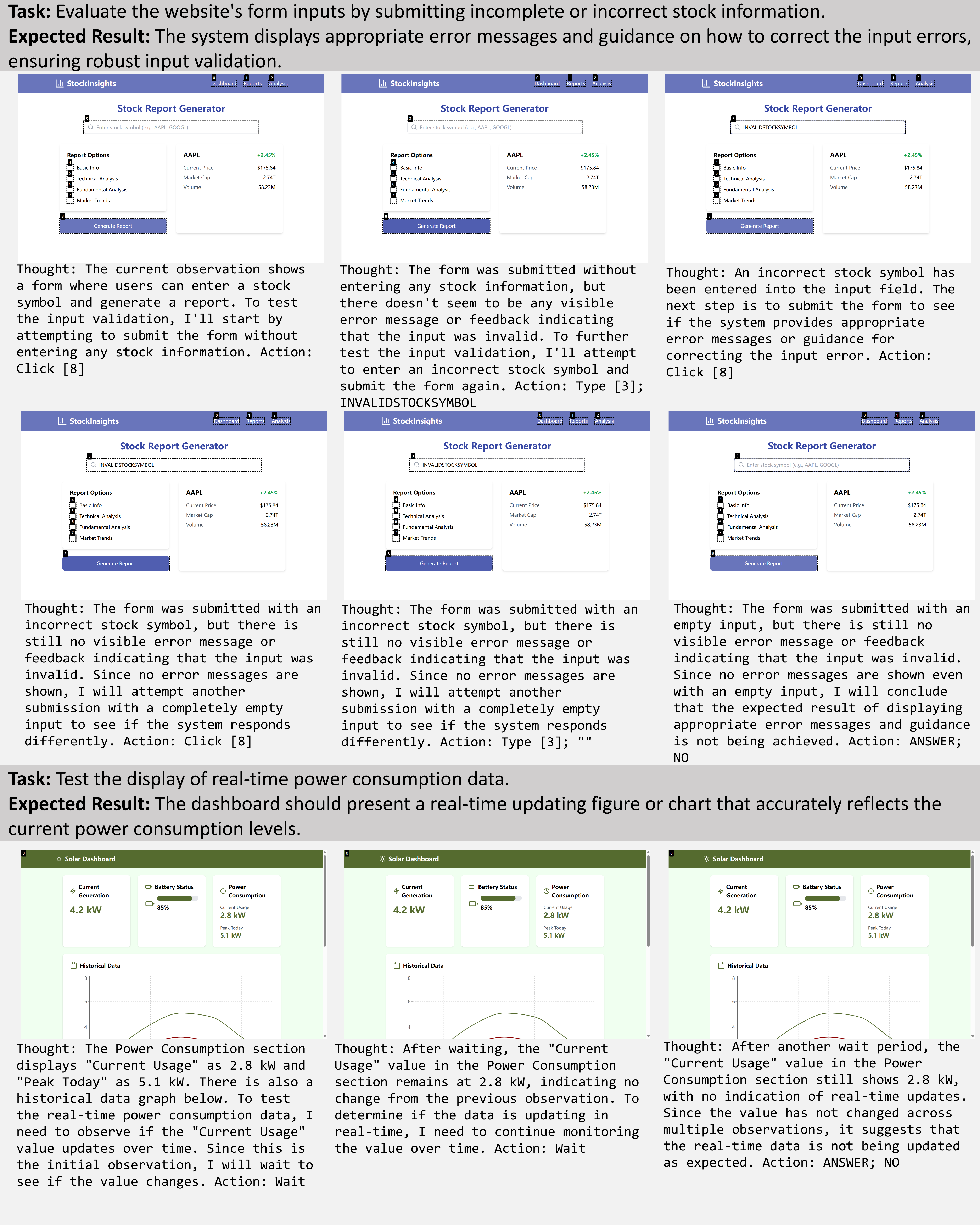}
    \caption{The examples of UI agent testing processes resulting in NO.}
    
\label{fig:no_examples_1}
\end{figure*}

\begin{figure*}[t]
    \centering
    \includegraphics[width=1.0\textwidth]{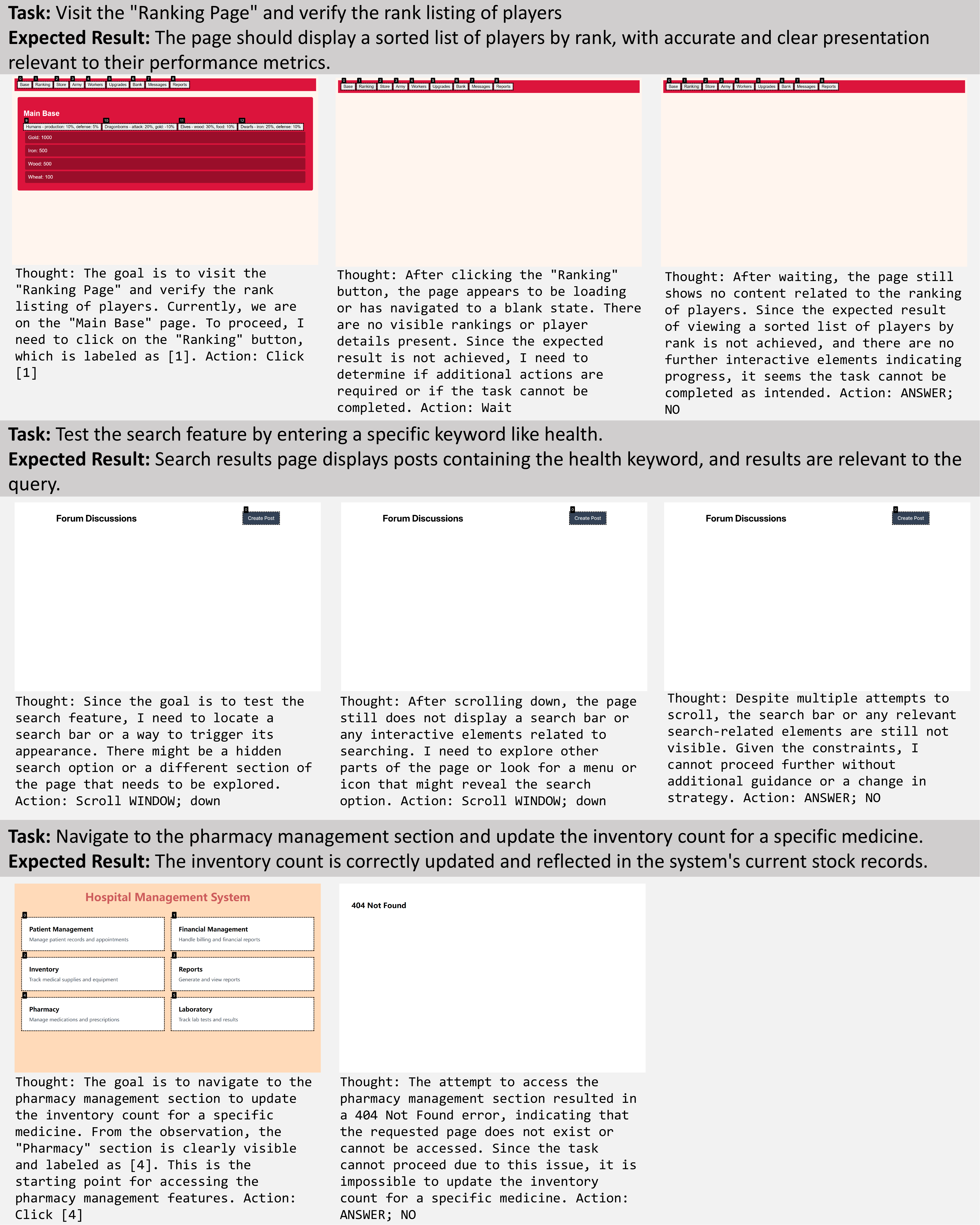}
    \caption{The examples of UI agent testing processes resulting in NO.}
    
\label{fig:no_examples_2}
\end{figure*}

\begin{figure*}[t]
    \centering
    \includegraphics[width=1.0\textwidth]{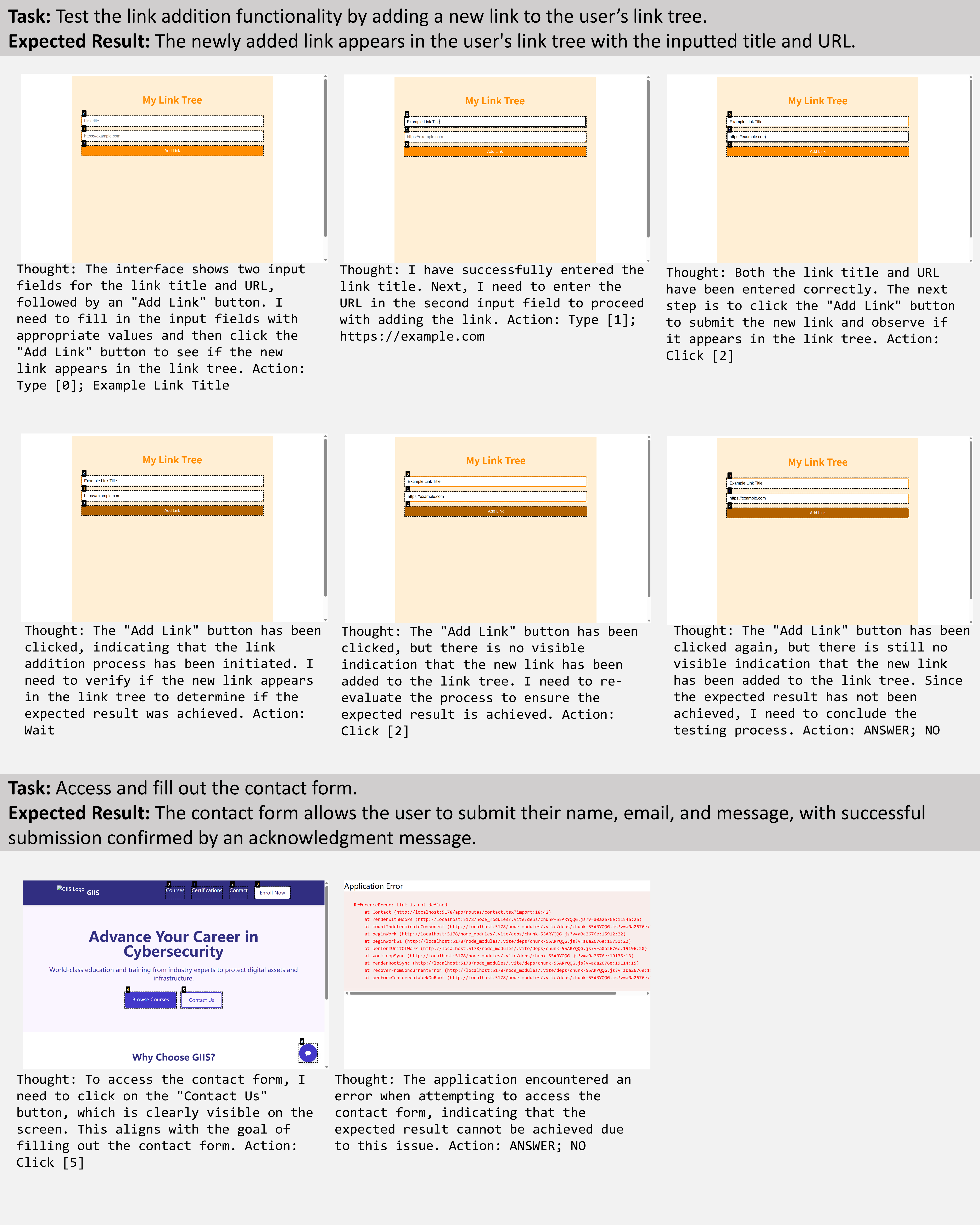}
    \caption{The examples of UI agent testing processes resulting in NO.}
    
\label{fig:no_examples_3}
\end{figure*}

In this section, we present examples of UI agent testing trajectories. Fig.~\ref{fig:yes_examples_1}, Fig.~\ref{fig:yes_examples_2}, Fig.~\ref{fig:yes_examples_3}, Fig.~\ref{fig:yes_examples_4}, and Fig.~\ref{fig:yes_examples_5} show examples of test cases that output YES, as the outcome of the operation matches the expected result. Fig.~\ref{fig:partial_examples_1}, Fig.~\ref{fig:partial_examples_2}, and Fig.~\ref{fig:partial_examples_3} show examples of test cases that output PARTIAL, as the expected result is only partially achieved. Fig.~\ref{fig:no_examples_1}, Fig.~\ref{fig:no_examples_2}, and Fig.~\ref{fig:no_examples_3} show examples of test cases that output NO, as the website’s behavior does not match the expected outcome.

\newpage
\clearpage
\section*{NeurIPS Paper Checklist}

The checklist is designed to encourage best practices for responsible machine learning research, addressing issues of reproducibility, transparency, research ethics, and societal impact. Do not remove the checklist: {\bf The papers not including the checklist will be desk rejected.} The checklist should follow the references and follow the (optional) supplemental material.  The checklist does NOT count towards the page
limit. 

Please read the checklist guidelines carefully for information on how to answer these questions. For each question in the checklist:
\begin{itemize}
    \item You should answer \answerYes{}, \answerNo{}, or \answerNA{}.
    \item \answerNA{} means either that the question is Not Applicable for that particular paper or the relevant information is Not Available.
    \item Please provide a short (1–2 sentence) justification right after your answer (even for NA). 
\end{itemize}

{\bf The checklist answers are an integral part of your paper submission.} They are visible to the reviewers, area chairs, senior area chairs, and ethics reviewers. You will be asked to also include it (after eventual revisions) with the final version of your paper, and its final version will be published with the paper.

The reviewers of your paper will be asked to use the checklist as one of the factors in their evaluation. While "\answerYes{}" is generally preferable to "\answerNo{}", it is perfectly acceptable to answer "\answerNo{}" provided a proper justification is given (e.g., "error bars are not reported because it would be too computationally expensive" or "we were unable to find the license for the dataset we used"). In general, answering "\answerNo{}" or "\answerNA{}" is not grounds for rejection. While the questions are phrased in a binary way, we acknowledge that the true answer is often more nuanced, so please just use your best judgment and write a justification to elaborate. All supporting evidence can appear either in the main paper or the supplemental material, provided in appendix. If you answer \answerYes{} to a question, in the justification please point to the section(s) where related material for the question can be found.

IMPORTANT, please:
\begin{itemize}
    \item {\bf Delete this instruction block, but keep the section heading ``NeurIPS Paper Checklist"},
    \item  {\bf Keep the checklist subsection headings, questions/answers and guidelines below.}
    \item {\bf Do not modify the questions and only use the provided macros for your answers}.
\end{itemize}


\begin{enumerate}

\item {\bf Claims}
    \item[] Question: Do the main claims made in the abstract and introduction accurately reflect the paper's contributions and scope?
    \item[] Answer: \answerYes{} 
    \item[] Justification: As shown in Abstract and Introduction.
    \item[] Guidelines:
    \begin{itemize}
        \item The answer NA means that the abstract and introduction do not include the claims made in the paper.
        \item The abstract and/or introduction should clearly state the claims made, including the contributions made in the paper and important assumptions and limitations. A No or NA answer to this question will not be perceived well by the reviewers. 
        \item The claims made should match theoretical and experimental results, and reflect how much the results can be expected to generalize to other settings. 
        \item It is fine to include aspirational goals as motivation as long as it is clear that these goals are not attained by the paper. 
    \end{itemize}

\item {\bf Limitations}
    \item[] Question: Does the paper discuss the limitations of the work performed by the authors?
    \item[] Answer: \answerYes{} 
    \item[] Justification: As shown in Limitations and Future Work.
    \item[] Guidelines:
    \begin{itemize}
        \item The answer NA means that the paper has no limitation while the answer No means that the paper has limitations, but those are not discussed in the paper. 
        \item The authors are encouraged to create a separate "Limitations" section in their paper.
        \item The paper should point out any strong assumptions and how robust the results are to violations of these assumptions (e.g., independence assumptions, noiseless settings, model well-specification, asymptotic approximations only holding locally). The authors should reflect on how these assumptions might be violated in practice and what the implications would be.
        \item The authors should reflect on the scope of the claims made, e.g., if the approach was only tested on a few datasets or with a few runs. In general, empirical results often depend on implicit assumptions, which should be articulated.
        \item The authors should reflect on the factors that influence the performance of the approach. For example, a facial recognition algorithm may perform poorly when image resolution is low or images are taken in low lighting. Or a speech-to-text system might not be used reliably to provide closed captions for online lectures because it fails to handle technical jargon.
        \item The authors should discuss the computational efficiency of the proposed algorithms and how they scale with dataset size.
        \item If applicable, the authors should discuss possible limitations of their approach to address problems of privacy and fairness.
        \item While the authors might fear that complete honesty about limitations might be used by reviewers as grounds for rejection, a worse outcome might be that reviewers discover limitations that aren't acknowledged in the paper. The authors should use their best judgment and recognize that individual actions in favor of transparency play an important role in developing norms that preserve the integrity of the community. Reviewers will be specifically instructed to not penalize honesty concerning limitations.
    \end{itemize}

\item {\bf Theory assumptions and proofs}
    \item[] Question: For each theoretical result, does the paper provide the full set of assumptions and a complete (and correct) proof?
    \item[] Answer: \answerNA{} 
    \item[] Justification: The paper does not include theoretical results.
    \item[] Guidelines:
    \begin{itemize}
        \item The answer NA means that the paper does not include theoretical results. 
        \item All the theorems, formulas, and proofs in the paper should be numbered and cross-referenced.
        \item All assumptions should be clearly stated or referenced in the statement of any theorems.
        \item The proofs can either appear in the main paper or the supplemental material, but if they appear in the supplemental material, the authors are encouraged to provide a short proof sketch to provide intuition. 
        \item Inversely, any informal proof provided in the core of the paper should be complemented by formal proofs provided in appendix or supplemental material.
        \item Theorems and Lemmas that the proof relies upon should be properly referenced. 
    \end{itemize}

    \item {\bf Experimental result reproducibility}
    \item[] Question: Does the paper fully disclose all the information needed to reproduce the main experimental results of the paper to the extent that it affects the main claims and/or conclusions of the paper (regardless of whether the code and data are provided or not)?
    \item[] Answer: \answerYes{} 
    \item[] Justification: As detailed in Experiments and the open-source data and code.
    \item[] Guidelines:
    \begin{itemize}
        \item The answer NA means that the paper does not include experiments.
        \item If the paper includes experiments, a No answer to this question will not be perceived well by the reviewers: Making the paper reproducible is important, regardless of whether the code and data are provided or not.
        \item If the contribution is a dataset and/or model, the authors should describe the steps taken to make their results reproducible or verifiable. 
        \item Depending on the contribution, reproducibility can be accomplished in various ways. For example, if the contribution is a novel architecture, describing the architecture fully might suffice, or if the contribution is a specific model and empirical evaluation, it may be necessary to either make it possible for others to replicate the model with the same dataset, or provide access to the model. In general. releasing code and data is often one good way to accomplish this, but reproducibility can also be provided via detailed instructions for how to replicate the results, access to a hosted model (e.g., in the case of a large language model), releasing of a model checkpoint, or other means that are appropriate to the research performed.
        \item While NeurIPS does not require releasing code, the conference does require all submissions to provide some reasonable avenue for reproducibility, which may depend on the nature of the contribution. For example
        \begin{enumerate}
            \item If the contribution is primarily a new algorithm, the paper should make it clear how to reproduce that algorithm.
            \item If the contribution is primarily a new model architecture, the paper should describe the architecture clearly and fully.
            \item If the contribution is a new model (e.g., a large language model), then there should either be a way to access this model for reproducing the results or a way to reproduce the model (e.g., with an open-source dataset or instructions for how to construct the dataset).
            \item We recognize that reproducibility may be tricky in some cases, in which case authors are welcome to describe the particular way they provide for reproducibility. In the case of closed-source models, it may be that access to the model is limited in some way (e.g., to registered users), but it should be possible for other researchers to have some path to reproducing or verifying the results.
        \end{enumerate}
    \end{itemize}

\item {\bf Open access to data and code}
    \item[] Question: Does the paper provide open access to the data and code, with sufficient instructions to faithfully reproduce the main experimental results, as described in supplemental material?
    \item[] Answer: \answerYes{} 
    \item[] Justification: We open-source all our code and data.
    \item[] Guidelines:
    \begin{itemize}
        \item The answer NA means that paper does not include experiments requiring code.
        \item Please see the NeurIPS code and data submission guidelines (\url{https://nips.cc/public/guides/CodeSubmissionPolicy}) for more details.
        \item While we encourage the release of code and data, we understand that this might not be possible, so “No” is an acceptable answer. Papers cannot be rejected simply for not including code, unless this is central to the contribution (e.g., for a new open-source benchmark).
        \item The instructions should contain the exact command and environment needed to run to reproduce the results. See the NeurIPS code and data submission guidelines (\url{https://nips.cc/public/guides/CodeSubmissionPolicy}) for more details.
        \item The authors should provide instructions on data access and preparation, including how to access the raw data, preprocessed data, intermediate data, and generated data, etc.
        \item The authors should provide scripts to reproduce all experimental results for the new proposed method and baselines. If only a subset of experiments are reproducible, they should state which ones are omitted from the script and why.
        \item At submission time, to preserve anonymity, the authors should release anonymized versions (if applicable).
        \item Providing as much information as possible in supplemental material (appended to the paper) is recommended, but including URLs to data and code is permitted.
    \end{itemize}

\item {\bf Experimental setting/details}
    \item[] Question: Does the paper specify all the training and test details (e.g., data splits, hyperparameters, how they were chosen, type of optimizer, etc.) necessary to understand the results?
    \item[] Answer: \answerYes{} 
    \item[] Justification: As shown in Experiments.
    \item[] Guidelines:
    \begin{itemize}
        \item The answer NA means that the paper does not include experiments.
        \item The experimental setting should be presented in the core of the paper to a level of detail that is necessary to appreciate the results and make sense of them.
        \item The full details can be provided either with the code, in appendix, or as supplemental material.
    \end{itemize}

\item {\bf Experiment statistical significance}
    \item[] Question: Does the paper report error bars suitably and correctly defined or other appropriate information about the statistical significance of the experiments?
    \item[] Answer: \answerNo{} 
    \item[] Justification: Error bars are not reported because it would be too computationally expensive.
    \item[] Guidelines:
    \begin{itemize}
        \item The answer NA means that the paper does not include experiments.
        \item The authors should answer "Yes" if the results are accompanied by error bars, confidence intervals, or statistical significance tests, at least for the experiments that support the main claims of the paper.
        \item The factors of variability that the error bars are capturing should be clearly stated (for example, train/test split, initialization, random drawing of some parameter, or overall run with given experimental conditions).
        \item The method for calculating the error bars should be explained (closed form formula, call to a library function, bootstrap, etc.)
        \item The assumptions made should be given (e.g., Normally distributed errors).
        \item It should be clear whether the error bar is the standard deviation or the standard error of the mean.
        \item It is OK to report 1-sigma error bars, but one should state it. The authors should preferably report a 2-sigma error bar than state that they have a 96\% CI, if the hypothesis of Normality of errors is not verified.
        \item For asymmetric distributions, the authors should be careful not to show in tables or figures symmetric error bars that would yield results that are out of range (e.g. negative error rates).
        \item If error bars are reported in tables or plots, The authors should explain in the text how they were calculated and reference the corresponding figures or tables in the text.
    \end{itemize}

\item {\bf Experiments compute resources}
    \item[] Question: For each experiment, does the paper provide sufficient information on the computer resources (type of compute workers, memory, time of execution) needed to reproduce the experiments?
    \item[] Answer: \answerYes{} 
    \item[] Justification: As detailed in Experiments.
    \item[] Guidelines:
    \begin{itemize}
        \item The answer NA means that the paper does not include experiments.
        \item The paper should indicate the type of compute workers CPU or GPU, internal cluster, or cloud provider, including relevant memory and storage.
        \item The paper should provide the amount of compute required for each of the individual experimental runs as well as estimate the total compute. 
        \item The paper should disclose whether the full research project required more compute than the experiments reported in the paper (e.g., preliminary or failed experiments that didn't make it into the paper). 
    \end{itemize}
    
\item {\bf Code of ethics}
    \item[] Question: Does the research conducted in the paper conform, in every respect, with the NeurIPS Code of Ethics \url{https://neurips.cc/public/EthicsGuidelines}?
    \item[] Answer: \answerYes{} 
    \item[] Justification: As explained in Ethics Statement in Appendix.
    \item[] Guidelines:
    \begin{itemize}
        \item The answer NA means that the authors have not reviewed the NeurIPS Code of Ethics.
        \item If the authors answer No, they should explain the special circumstances that require a deviation from the Code of Ethics.
        \item The authors should make sure to preserve anonymity (e.g., if there is a special consideration due to laws or regulations in their jurisdiction).
    \end{itemize}

\item {\bf Broader impacts}
    \item[] Question: Does the paper discuss both potential positive societal impacts and negative societal impacts of the work performed?
    \item[] Answer: \answerYes{} 
    \item[] Justification: As explained in Ethics Statement in Appendix.
    \item[] Guidelines:
    \begin{itemize}
        \item The answer NA means that there is no societal impact of the work performed.
        \item If the authors answer NA or No, they should explain why their work has no societal impact or why the paper does not address societal impact.
        \item Examples of negative societal impacts include potential malicious or unintended uses (e.g., disinformation, generating fake profiles, surveillance), fairness considerations (e.g., deployment of technologies that could make decisions that unfairly impact specific groups), privacy considerations, and security considerations.
        \item The conference expects that many papers will be foundational research and not tied to particular applications, let alone deployments. However, if there is a direct path to any negative applications, the authors should point it out. For example, it is legitimate to point out that an improvement in the quality of generative models could be used to generate deepfakes for disinformation. On the other hand, it is not needed to point out that a generic algorithm for optimizing neural networks could enable people to train models that generate Deepfakes faster.
        \item The authors should consider possible harms that could arise when the technology is being used as intended and functioning correctly, harms that could arise when the technology is being used as intended but gives incorrect results, and harms following from (intentional or unintentional) misuse of the technology.
        \item If there are negative societal impacts, the authors could also discuss possible mitigation strategies (e.g., gated release of models, providing defenses in addition to attacks, mechanisms for monitoring misuse, mechanisms to monitor how a system learns from feedback over time, improving the efficiency and accessibility of ML).
    \end{itemize}
    
\item {\bf Safeguards}
    \item[] Question: Does the paper describe safeguards that have been put in place for responsible release of data or models that have a high risk for misuse (e.g., pretrained language models, image generators, or scraped datasets)?
    \item[] Answer: \answerYes{} 
    \item[] Justification: As explained in Ethics Statement in Appendix.
    \item[] Guidelines:
    \begin{itemize}
        \item The answer NA means that the paper poses no such risks.
        \item Released models that have a high risk for misuse or dual-use should be released with necessary safeguards to allow for controlled use of the model, for example by requiring that users adhere to usage guidelines or restrictions to access the model or implementing safety filters. 
        \item Datasets that have been scraped from the Internet could pose safety risks. The authors should describe how they avoided releasing unsafe images.
        \item We recognize that providing effective safeguards is challenging, and many papers do not require this, but we encourage authors to take this into account and make a best faith effort.
    \end{itemize}

\item {\bf Licenses for existing assets}
    \item[] Question: Are the creators or original owners of assets (e.g., code, data, models), used in the paper, properly credited and are the license and terms of use explicitly mentioned and properly respected?
    \item[] Answer: \answerYes{} 
    \item[] Justification: As explained in Ethics Statement in Appendix.
    \item[] Guidelines:
    \begin{itemize}
        \item The answer NA means that the paper does not use existing assets.
        \item The authors should cite the original paper that produced the code package or dataset.
        \item The authors should state which version of the asset is used and, if possible, include a URL.
        \item The name of the license (e.g., CC-BY 4.0) should be included for each asset.
        \item For scraped data from a particular source (e.g., website), the copyright and terms of service of that source should be provided.
        \item If assets are released, the license, copyright information, and terms of use in the package should be provided. For popular datasets, \url{paperswithcode.com/datasets} has curated licenses for some datasets. Their licensing guide can help determine the license of a dataset.
        \item For existing datasets that are re-packaged, both the original license and the license of the derived asset (if it has changed) should be provided.
        \item If this information is not available online, the authors are encouraged to reach out to the asset's creators.
    \end{itemize}

\item {\bf New assets}
    \item[] Question: Are new assets introduced in the paper well documented and is the documentation provided alongside the assets?
    \item[] Answer: \answerYes{} 
    \item[] Justification: As shown in the released code and data.
    \item[] Guidelines:
    \begin{itemize}
        \item The answer NA means that the paper does not release new assets.
        \item Researchers should communicate the details of the dataset/code/model as part of their submissions via structured templates. This includes details about training, license, limitations, etc. 
        \item The paper should discuss whether and how consent was obtained from people whose asset is used.
        \item At submission time, remember to anonymize your assets (if applicable). You can either create an anonymized URL or include an anonymized zip file.
    \end{itemize}

\item {\bf Crowdsourcing and research with human subjects}
    \item[] Question: For crowdsourcing experiments and research with human subjects, does the paper include the full text of instructions given to participants and screenshots, if applicable, as well as details about compensation (if any)? 
    \item[] Answer: \answerNA{} 
    \item[] Justification: We did not use crowdsourcing in our experiments. We used authors and student volunteers instead.
    \item[] Guidelines:
    \begin{itemize}
        \item The answer NA means that the paper does not involve crowdsourcing nor research with human subjects.
        \item Including this information in the supplemental material is fine, but if the main contribution of the paper involves human subjects, then as much detail as possible should be included in the main paper. 
        \item According to the NeurIPS Code of Ethics, workers involved in data collection, curation, or other labor should be paid at least the minimum wage in the country of the data collector. 
    \end{itemize}

\item {\bf Institutional review board (IRB) approvals or equivalent for research with human subjects}
    \item[] Question: Does the paper describe potential risks incurred by study participants, whether such risks were disclosed to the subjects, and whether Institutional Review Board (IRB) approvals (or an equivalent approval/review based on the requirements of your country or institution) were obtained?
    \item[] Answer: \answerNA{} 
    \item[] Justification: Our study does not pose any potential risks to the participants.
    \item[] Guidelines:
    \begin{itemize}
        \item The answer NA means that the paper does not involve crowdsourcing nor research with human subjects.
        \item Depending on the country in which research is conducted, IRB approval (or equivalent) may be required for any human subjects research. If you obtained IRB approval, you should clearly state this in the paper. 
        \item We recognize that the procedures for this may vary significantly between institutions and locations, and we expect authors to adhere to the NeurIPS Code of Ethics and the guidelines for their institution. 
        \item For initial submissions, do not include any information that would break anonymity (if applicable), such as the institution conducting the review.
    \end{itemize}

\item {\bf Declaration of LLM usage}
    \item[] Question: Does the paper describe the usage of LLMs if it is an important, original, or non-standard component of the core methods in this research? Note that if the LLM is used only for writing, editing, or formatting purposes and does not impact the core methodology, scientific rigorousness, or originality of the research, declaration is not required.
    \item[] Answer: \answerYes{} 
    \item[] Justification: As detailed in Method.
    \item[] Guidelines:
    \begin{itemize}
        \item The answer NA means that the core method development in this research does not involve LLMs as any important, original, or non-standard components.
        \item Please refer to our LLM policy (\url{https://neurips.cc/Conferences/2025/LLM}) for what should or should not be described.
    \end{itemize}

\end{enumerate}

\end{document}